\documentclass[final,5p,times,twocolumn,sort&compress]{elsarticle}

\usepackage{setspace}
\usepackage{algorithm}
\usepackage{algorithmicx}
\usepackage{algpseudocode}
\usepackage{amsmath}
\usepackage{amsthm}
\usepackage{framed}
\usepackage{xspace}
\usepackage{warpcol}

\usepackage{graphicx}
\usepackage{epsfig}
\usepackage{subfigure}
\usepackage{dcolumn}
\usepackage{float}
\usepackage[normalem]{ulem}
\useunder{\uline}{\ul}{}
\usepackage{multirow}
\usepackage{listings}
\usepackage{booktabs}

\usepackage{mathtools, amssymb}

\makeatletter 
\renewcommand{\@thesubfigure}{\hskip\subfiglabelskip}
 \makeatother

\journal{arxiv}

\begin{document}

\begin{frontmatter}



\title{Joint Object Contour Points and Semantics for Instance Segmentation}

\author[label1]{Wenchao Zhang}
\author{Chong Fu\fnref{label1,label2,label3} \corref{cor1}}
\ead{fuchong@mail.neu.edu.cn}
\cortext[cor1]{Corresponding author.}
\author[label1]{Mai Zhu}
\author[label4]{Lin Cao}
\author[label5]{Ming Tie}
\author[label6]{Chiu-Wing Sham}

\address[label1]{School of Computer Science and Engineering, Northeastern University, 
  Shenyang 110819, China}
\address[label2]{Engineering Research Center of Security Technology of Complex Network System, Ministry of Education, China}
\address[label3]{Key Laboratory of Intelligent Computing in Medical Image, Ministry of Education, Northeastern University, Shenyang 110819, China}
\address[label4]{School of Information and Communication Engineering, Beijing Information Science and Technology University, Beijing 100101, China}
\address[label5]{Science and Technology on Space Physics Laboratory, Beijing 100076, China}
\address[label6]{School of Computer Science, The University of Auckland, New Zealand}

\begin{abstract}
The attributes of object contours has great significance for instance segmentation task. However, most of the current popular deep neural networks do not pay much attention to the object edge information. Inspired by the human annotation process when making instance segmentation datasets, in this paper, we propose Mask Point R-CNN aiming at promoting the neural network's attention to the object boundary. Specifically, we innovatively extend the original human keypoint detection task to the contour point detection of any object. Based on this analogy, we present an contour point detection auxiliary task to Mask R-CNN, which can boost the gradient flow between different tasks by effectively using feature fusion strategies and multi-task joint training. As a consequence, the model will be more sensitive to the edges of the object and can capture more geometric features. Quantitatively, the experimental results show that our approach outperforms vanilla Mask R-CNN by 3.8\% on Cityscapes dataset and 0.8\% on COCO dataset.

\end{abstract}




\end{frontmatter}


\section{Introduction}
{I}{n} recent years, due to the emergence of deep networks, the performance of many computer vision applications has been dramatically improved. Classification, object detection, and instances segmentation are hot topics in computer vision community, especially instances segmentation, which includes two tasks: instance recognition and semantic segmentation. Compared with object detection and image level classification, it requires more sophisticated annotation information, therefore, its framework requires a more complex design.

Recently, like Mask R-CNN~\cite{he2017mask} and MaskLab~\cite{chen2018masklab}, the champion algorithms of COCO dataset~\cite{lin2014microsoft} instance segmentation task, the main idea is to divide the instance segmentation pipeline into two stages: the first stage is to locate and classify the objects in the image, and the second stage is to classify the foreground in the bounding box. In this paper, we focus on improving the instance segmentation task of Mask R-CNN, which is a further extension of Faster R-CNN~\cite{ren2015faster}. Specifically, it adds a parallel mask head to the original Faster R-CNN box head. First of all, it obtains the region of interest with a spatial scale of $14\times14$ through RoIAlign operation, and then a mask tensor with a spatial scale of $28\times28$ is predicted by a small head network with a series of convolution and upsampling operations. But this kind of prediction result is only obtained through binary classification and thereby cannot well represent the geometric properties of the object. To address this drawback, we added an auxiliary task on the basis of Mask R-CNN. As is known, in the pipeline of Mask R-CNN, for the person category in COCO dataset, it can not only perform detection and instance segmentation, alternatively, it can also perform human keypoint detection task simultaneously (i.e. keypoint R-CNN). Inspired by the process of labeling instance segmentation objects, we innovatively extend the keypoint detection task to all object categories. In other words, we add a new keypoint detection auxiliary task to Mask R-CNN, which can enhance the model's attention to the object boundary by detecting the contour of the object. To sum up, in our method, we use mask head pixel-by-pixel classification as the main task, and focus on the object edge information by aggregating the target edge points simutanously. As a consequence, this multi-task joint training model can well transform the geometric features of the object into the constraints of the binary mask of instance segmentation, and further sharpen the edges of the instance segmentation results to improve the detection performance. Without losing generality, we validate our method on the Cityscapes dataset~\cite{cordts2016cityscapes} and the large-scale object detection COCO dataset, and compared with Mask R-CNN baseline with ResNet-50, our method can effectively improve the performance of the model, the AP (Average precision) value on the Cityscapes dataset is 36.9(ours) vs 33.1, and the result on the COCO dataset is 35.4(ours) vs 34.6.

\begin{figure*}
\begin{center}
\includegraphics[width=16cm,height=8.83cm]{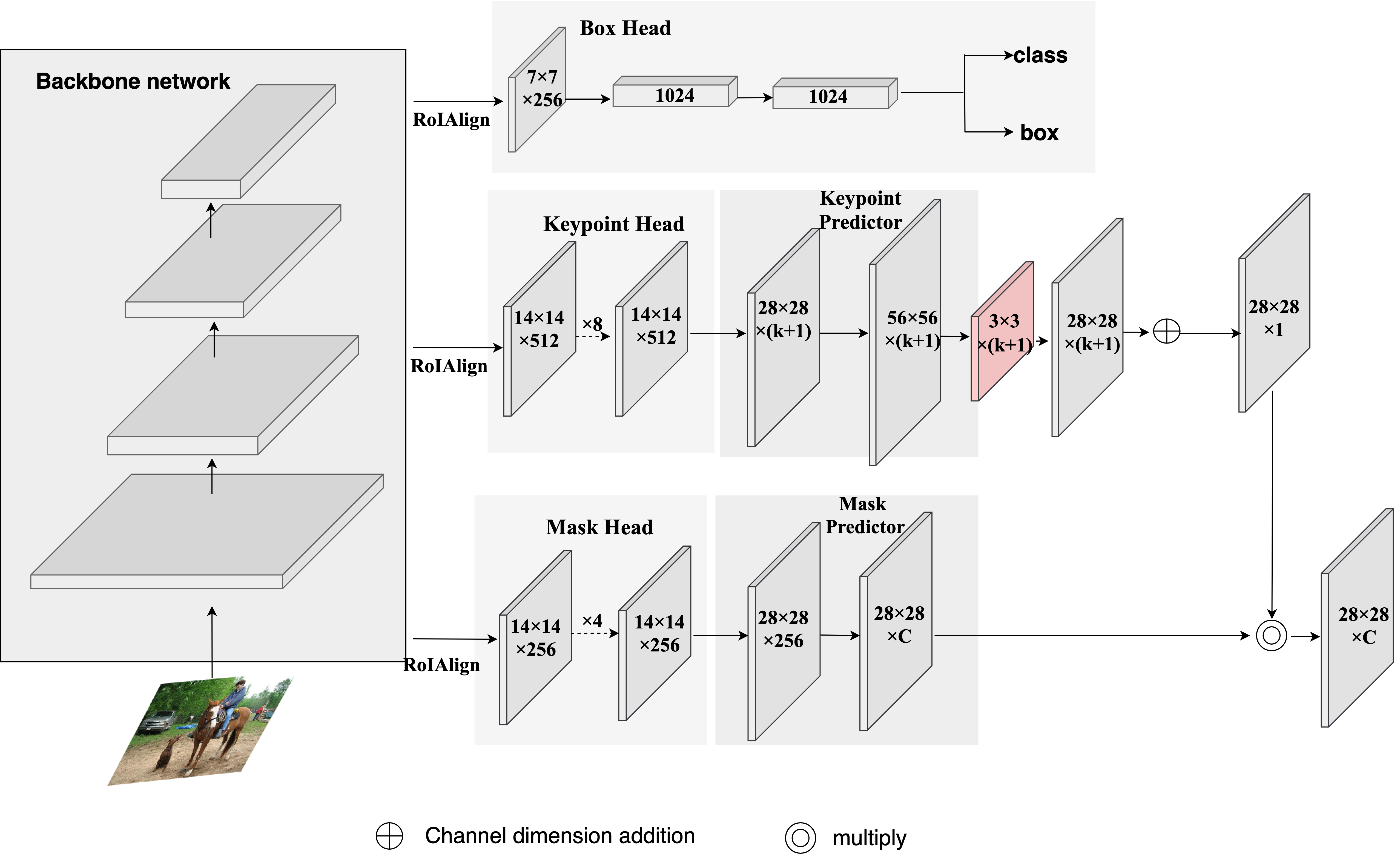}
\end{center}{}
   \caption{Network architecture of Mask Point R-CNN. In our model, the backbone network is in charge of feature extraction and we add a new keypoint head to the box head and mask head of mask rcnn, which is used to detect the boundary points of objects.}
\label{fig:short}
\label{fig:fig1}
\end{figure*}

In summary, the main contributions of this work are highlighted as follows:

  1. We propose the  Mask Point R-CNN, illustrated in Fig. ~\ref{fig:fig1}, which is a new exploration of instance segmentation by combining the edge information of the object.
  
  2. Compared to Mask R-CNN, we utilize features in the mask prediction branch and keypoint prediction branch with different attributes, and perform joint training through multi-tasks features fusion. 

  3. Different from the human body keypoint detection, we extend the keypoint detection task to any category boundary point detection, and experimentally prove that it can enhance the network's attention to the model boundary.

The rest of this paper is organized as follows. In Section \ref{related}, we briefly review related work on instance segmentation, multi-task learning, and keypoint detection. In Section \ref{method}, we describe the motivation and details of our proposed algorithm. Experimental details and analysis of the results are elaborated in Section \ref{expe}. In Section \ref{futurework}, we analyze the interaction between our method subtasks and verified their effectiveness through visualization. Finally, we conclude the paper in Section \ref{Conclusion}.
\section{Related Work}
\label{related}

This section provides an overview of literatures on instance segmentation, multi-task learning, and keypoint detection.

\subsection{Instance segmentation}
There are two common ways of achieving instance segmentation: detection first and segmentation first. The detection first method consists of using the object detector to obtain the bounding boxes coordinates preferentially and then segment the instances in the bounding boxes. As an outstanding representative work, He $et~al$.~\cite{he2017mask} proposed Mask R-CNN to perform the instance segmentation task by adding a mask prediction branch on the basis of object classification and bounding boxes regression task of Faster R-CNN~\cite{ren2015faster}. Many subsequent methods were proposed based on Mask R-CNN~\cite{huang2019mask,liu2018path,chen2019hybrid,qiao2020detectors,chen2018masklab}. For instance, Chen $et~al$.~\cite{chen2018masklab}proposed MaskLab that utilized semantic segmentation and direction prediction to implement mask prediction. Huang $et~al$.~\cite{huang2019mask} proposed MS R-CNN(Mask Scoring R-CNN) that added a mask IoU (Intersection over Union) head by combining instance features and corresponding prediction masks in Mask R-CNN to enhance the consistency between mask quality and mask score. Liu $et~al$.~\cite{liu2018path} proposed PAN (Path Aggregation Network) which combined more low-level features through Bottom-up Path Augmentation, and obtained more accurate segmentation through AFP (Adaptive Feature Pooling) and fully-connected fusion. Recently, there are also many excellent detection first methods that focus on building real-time instance segmentation systems~\cite{xie2020polarmask,bolya2019yolact,bolya2019yolact++,chen2020blendmask,lee2020centermask,peng2020deep}. Contrary to the previous method, the segmentation first method is to classify each pixel in the image preferentially and then group them into a single instance. For instance, Hariharan $et~al$. ~\cite{hariharan2014simultaneous} proposed SDS (Simultaneous detection and segmentation) to locate segmentation candidate regions. This method uses multi-scale combinatorial grouping ~\cite{arbelaez2014multiscale} and obtain bounding boxes simultaneously to achieve instance segmentation. The InstanceCut~\cite{kirillov2017instancecut} method utilized the edge graph to divide the segmentation map into different objects. Analogously, the work aforementioned is to perform segmentation tasks by classifying the foreground and background, and our work is to get a better instance segmentation effect by paying attention to the boundary of the object.

\subsection{Multi-Task learning}
 MTL (Multi-Task Learning) is an inductive transfer method that makes full use of domain-specific information hidden in the training signals of multiple related tasks. During backward propagation, multi-task learning allows shared hidden layer-specific features to be used by other tasks. In particular, MTL can learn features that are applicable to several different tasks. Such features are often not easy to learn in a single-task learning network. MTL is widely used in computer vision~\cite{huang2019mask,chen2019hybrid,qiao2020detectors}, speech, natural language processing, and other fields~\cite{zhao2017feature}. In~\cite{su2015multi}, Su $et~al$. calibrates missing features in the learning process by combining low-level raw binary attributes and intermediate attributes. In~\cite{abdulnabi2015multi}, Abdulnabi $et~al$. proposed the use of a combination of hidden matrix and multi-matrix to decouple model parameters to allow CNN (Convolutional Neural Network) models to share visual knowledge between different categories simultaneously. In~\cite{yuan2013multi}, Yuan $et~al$. enhanced the robustness of model coefficient estimation by constructing a hierarchical sparse learning model of multi-task interconnection. In~\cite{yim2015rotating}, Yim $et~al$. added an additional task to improve the ability of DNN (Deep Neural Network) to maintain face identity, which using a human face and a binary target pose code to generate a face image with the same identity and target pose. In our method, we use the keypoint detection task in conjunction with the instance segmentation task in the original Mask R-CNN to calibrate the boundary features of the object, the purpose of which is to enhance the gradient flow between different tasks and the fusion of different attribute features.

\subsection{Keypoint detection}
  Keypoint detection technology has been widely used in computer vision research, such as human pose detection tasks~\cite{pishchulin2016deepcut, xiao2018simple, osokin2018real} and anchor free object detection tasks~\cite{zhou2019objects, zhou2019bottom, law2018cornernet, maninis2018deep}. The purpose of human body keypoint estimation task is to detect $k$ 2D keypoints of the human body, such as joints, facial features, etc., and describes human bone information by these keypoints. Lately, keypoint estimation techniques are also used in many detection tasks. Generally, the current object detection method can be divided into anchor base~\cite{redmon2016you, redmon2017yolo9000, liu2016ssd, redmon2018, girshick2015fast, ren2015faster} and anchor free according to whether the proposal is obtained by sliding window. Specifically, the anchor base method utilizes sliding window to generate a bunch of candidate anchor boxes, and then remove the redundant bounding boxes by NMS (Non-Maximum Suppression)~\cite{bodla2017soft} to get the final detection results. The anchor free method utilizes keypoint detection technology~\cite{yu2018deep, newell2016stacked} to obtain the center point or extreme point of the object for target positioning. By contrast, the detection points of this method are all located in the interior of the detected object, so it is more conducive to the learning of network features. Following that intuition, we utilize the keypoint detection branch to detect the edge aggregation points of the target to capture more geometric features of objects.

\section{Method}
\label{method}
In this section, we will first introduce the motivation of our method, and then briefly review the Mask R-CNN algorithm, finally, describe the design of our model in detail from three aspects: boundary point extraction, model framework, and loss function.

\begin{figure*}[!htb]
  \centering
    \subfigbottomskip=0.1pt
    \subfigure{\includegraphics[width=0.22\linewidth]{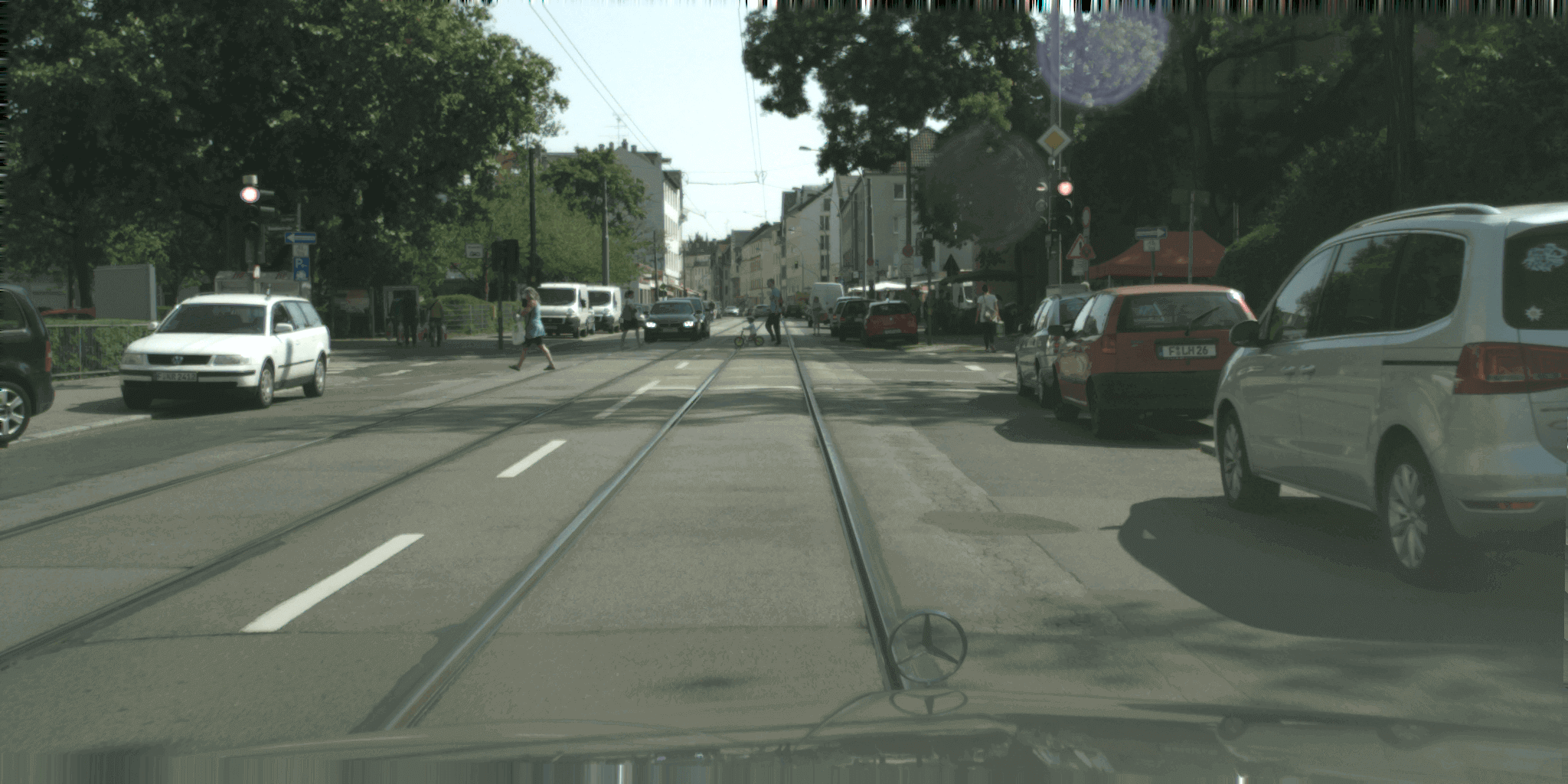}}  
  \subfigure{\includegraphics[width=0.22\linewidth]{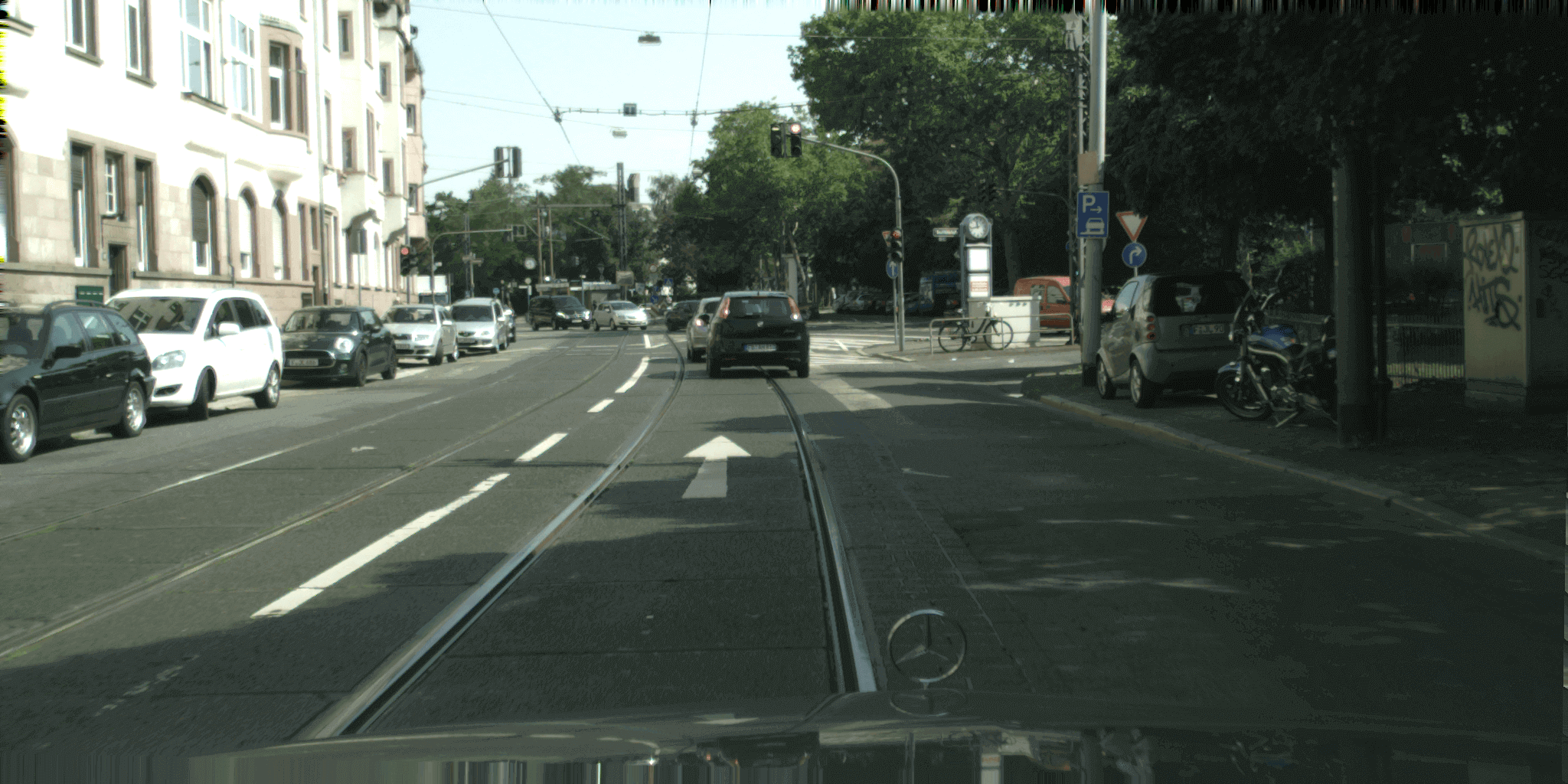}} 
    \subfigure{\includegraphics[width=0.22\linewidth]{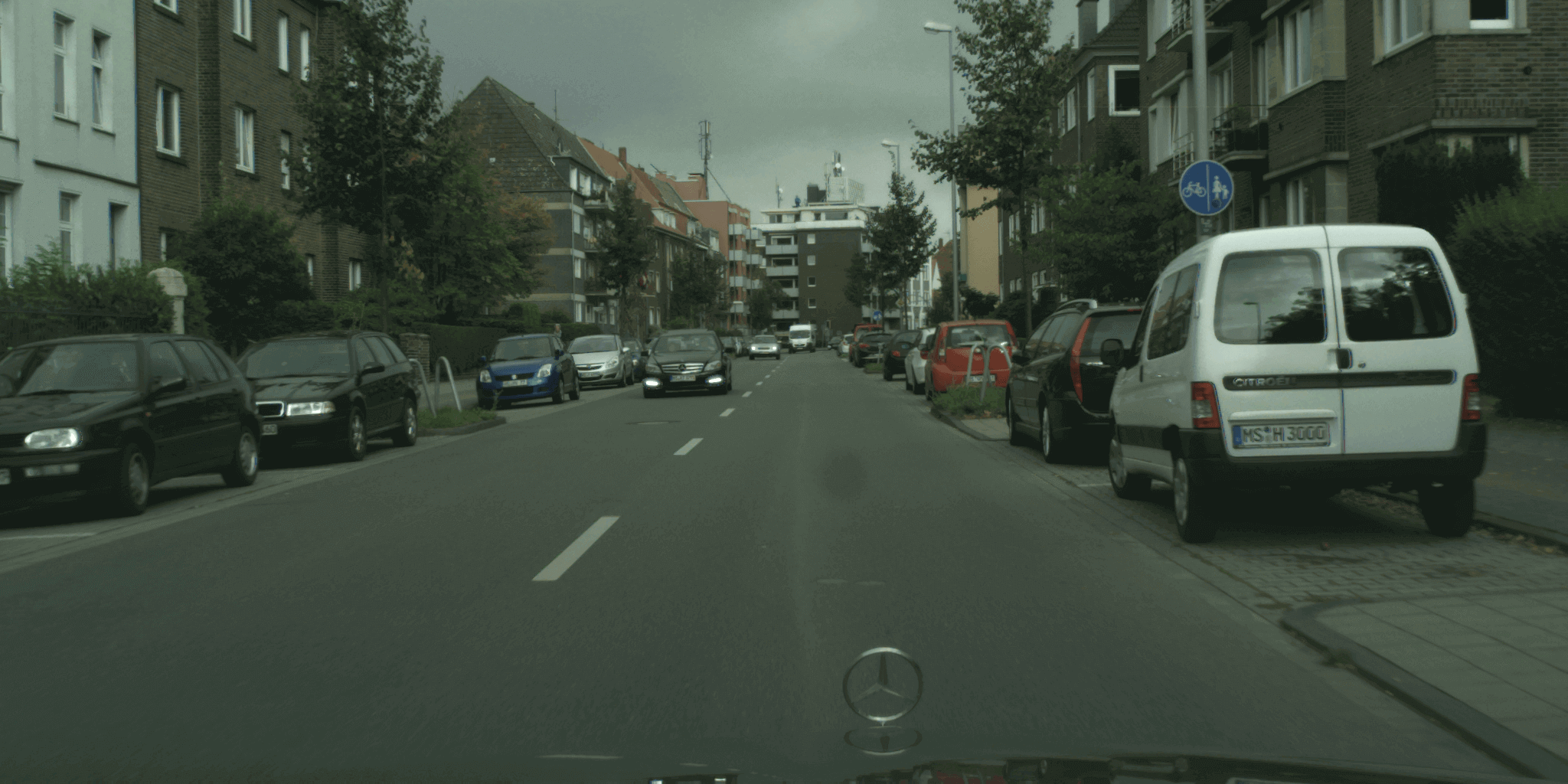}} 
    \subfigure{\includegraphics[width=0.22\linewidth]{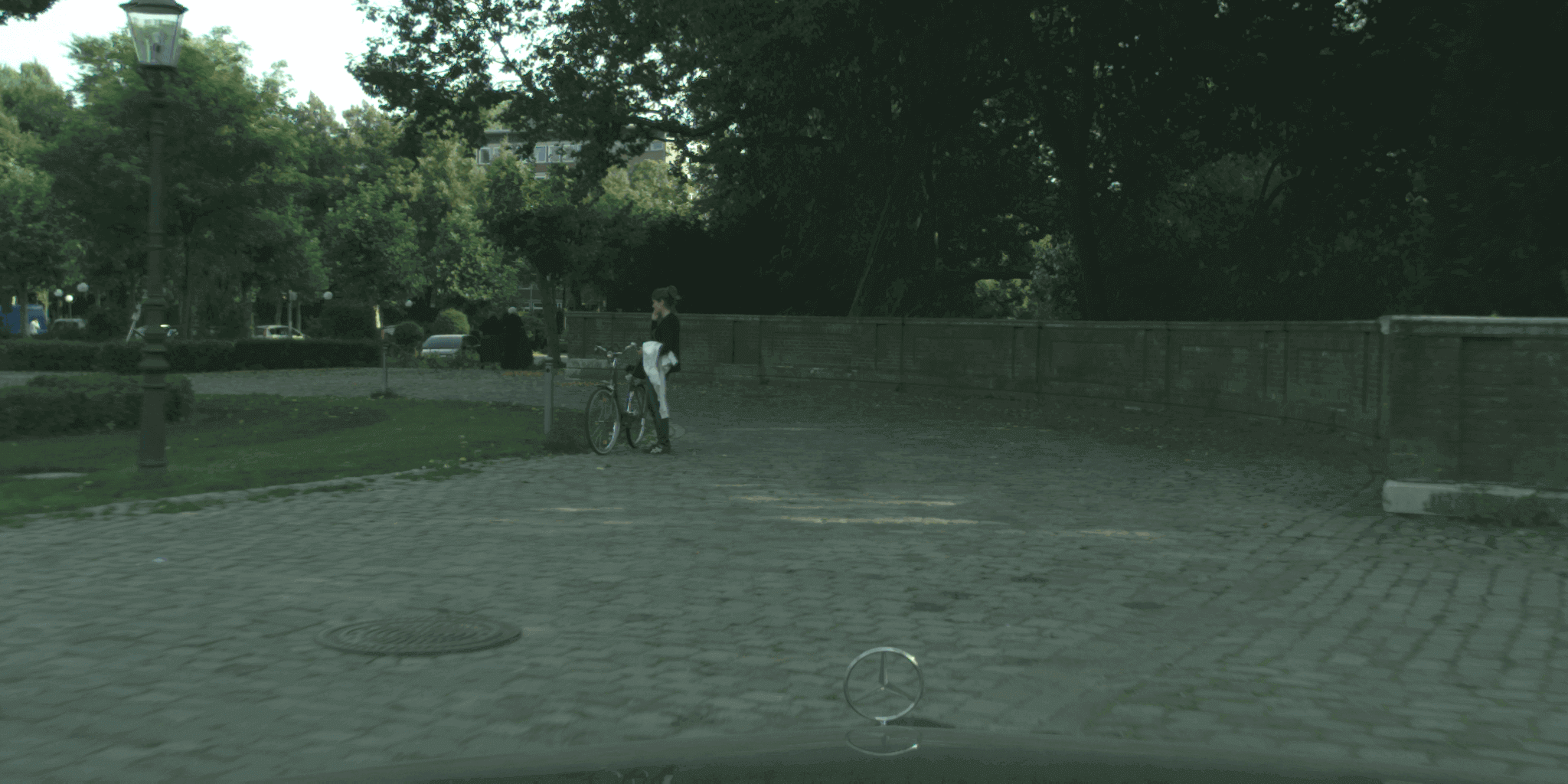}} \\
  \subfigure{\includegraphics[width=0.22\linewidth]{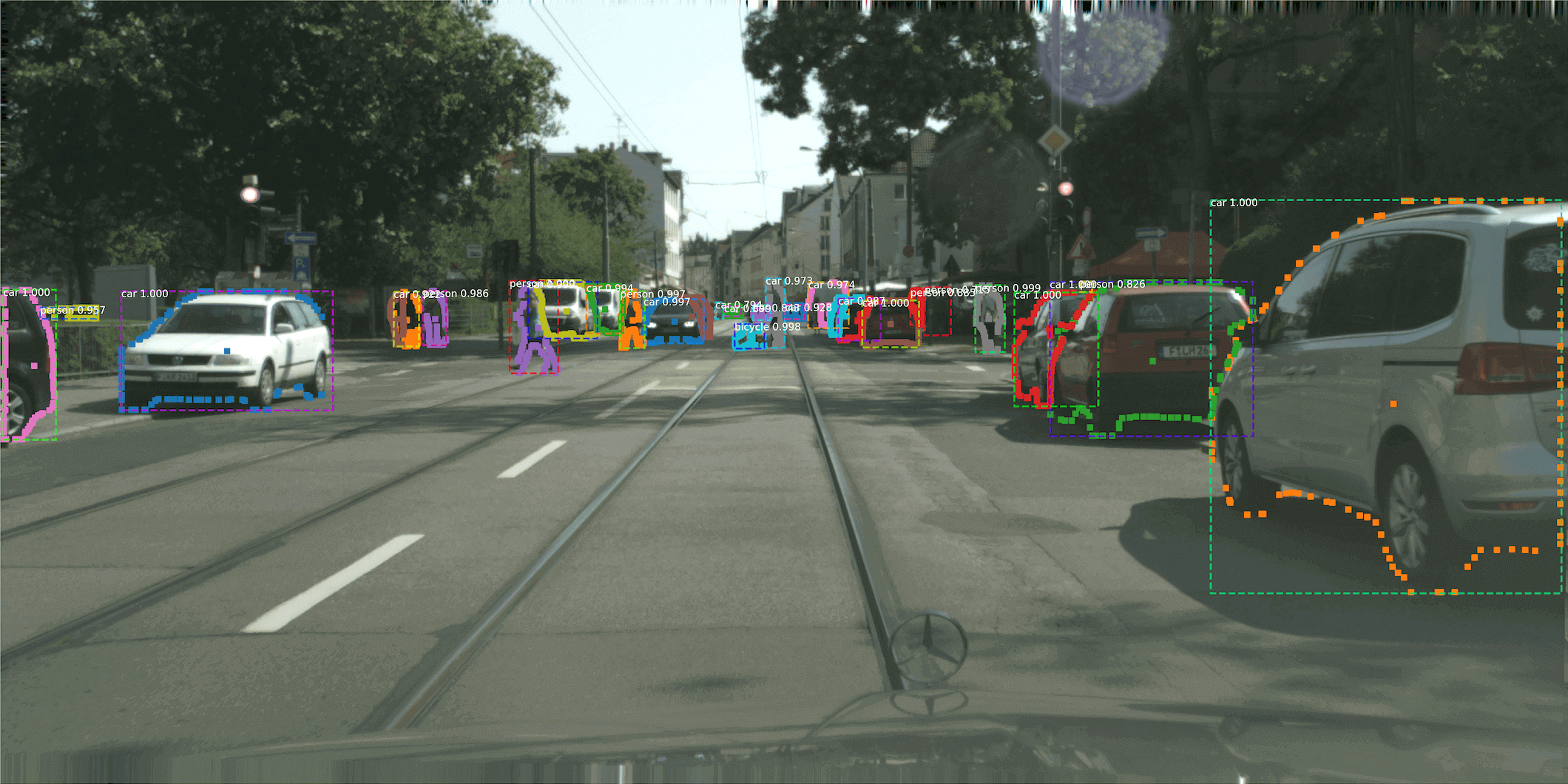}} 
    \subfigure{\includegraphics[width=0.22\linewidth]{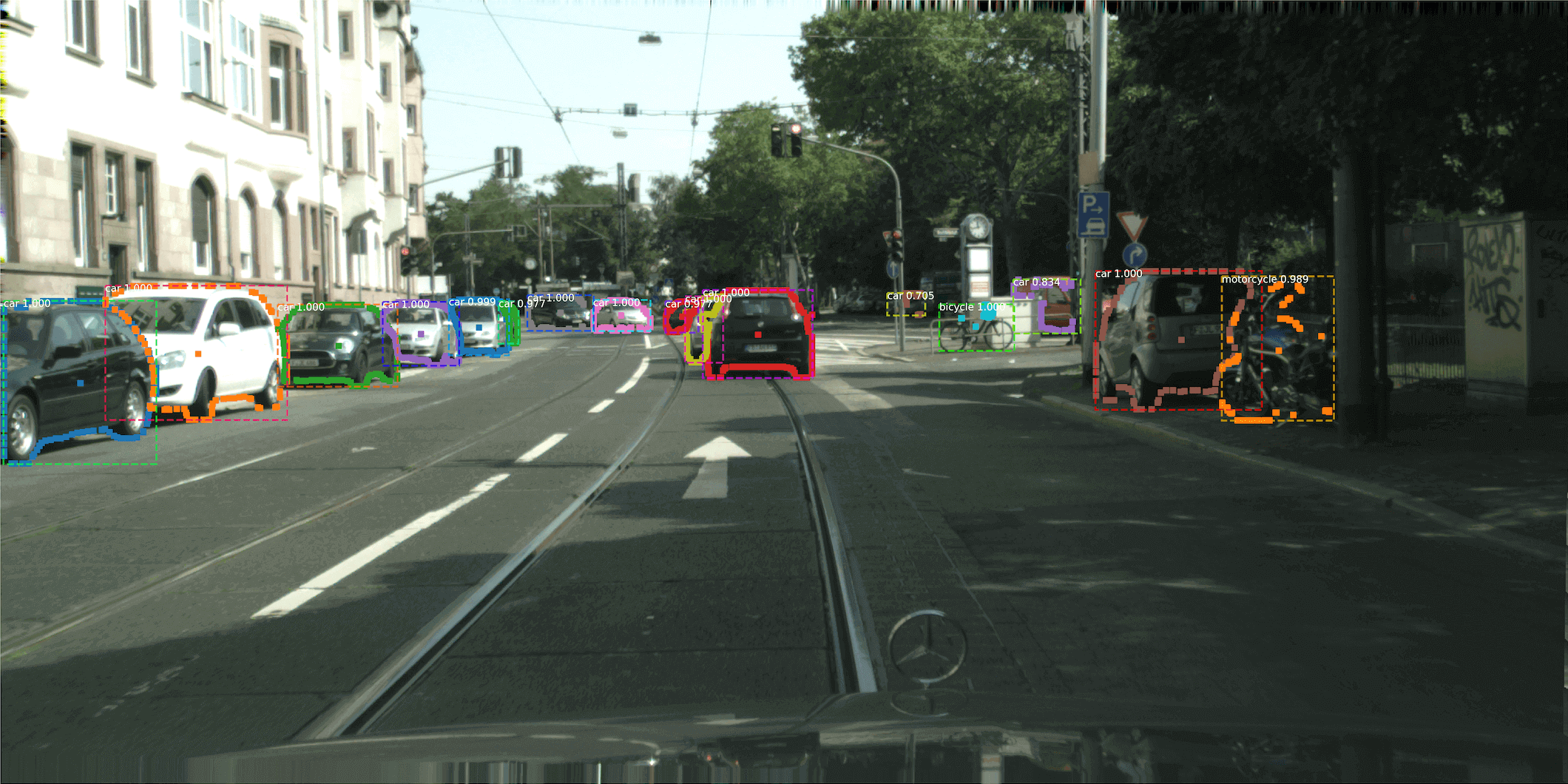}} 
    \subfigure{\includegraphics[width=0.22\linewidth]{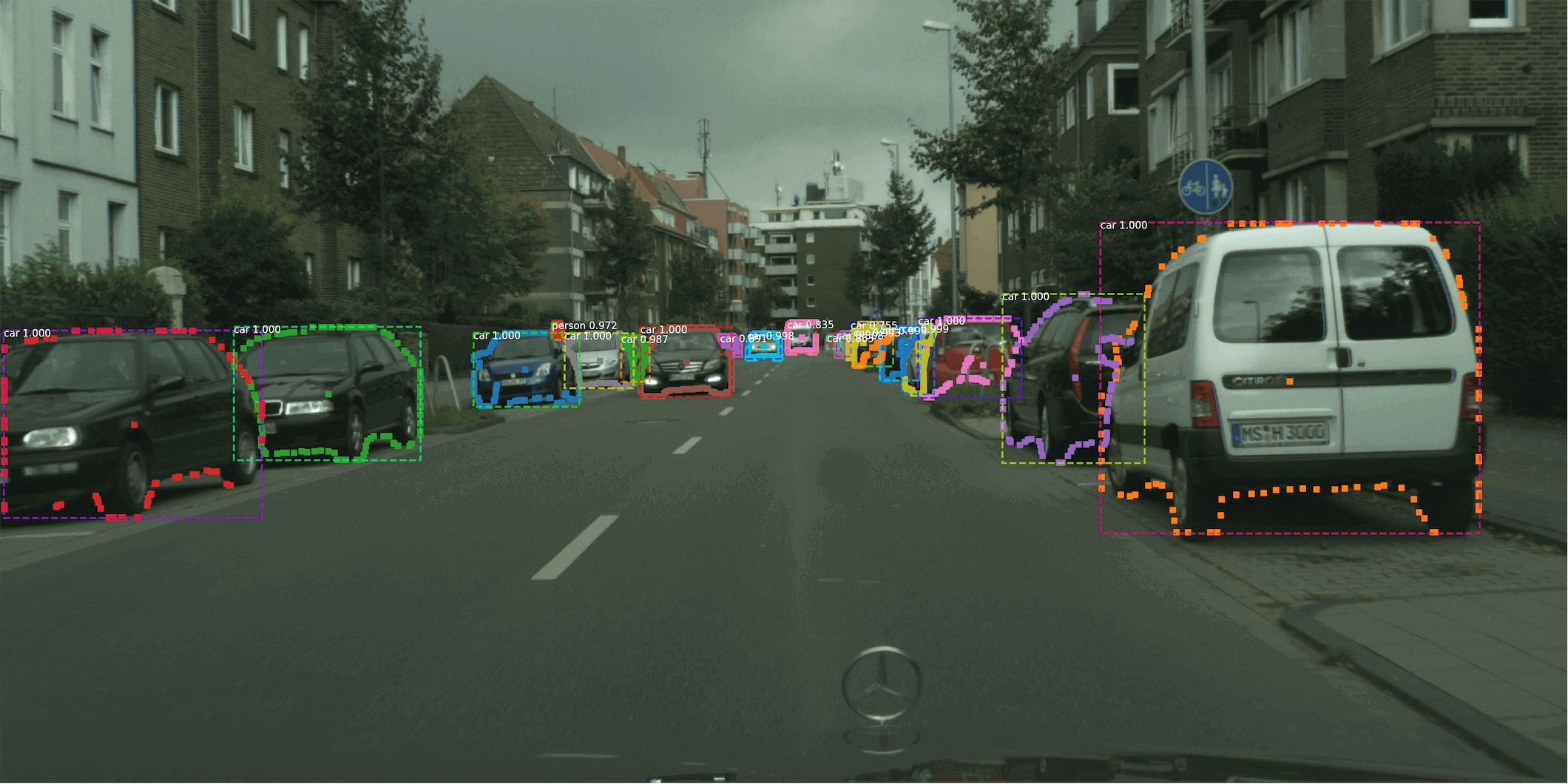}}  
  \subfigure{\includegraphics[width=0.22\linewidth]{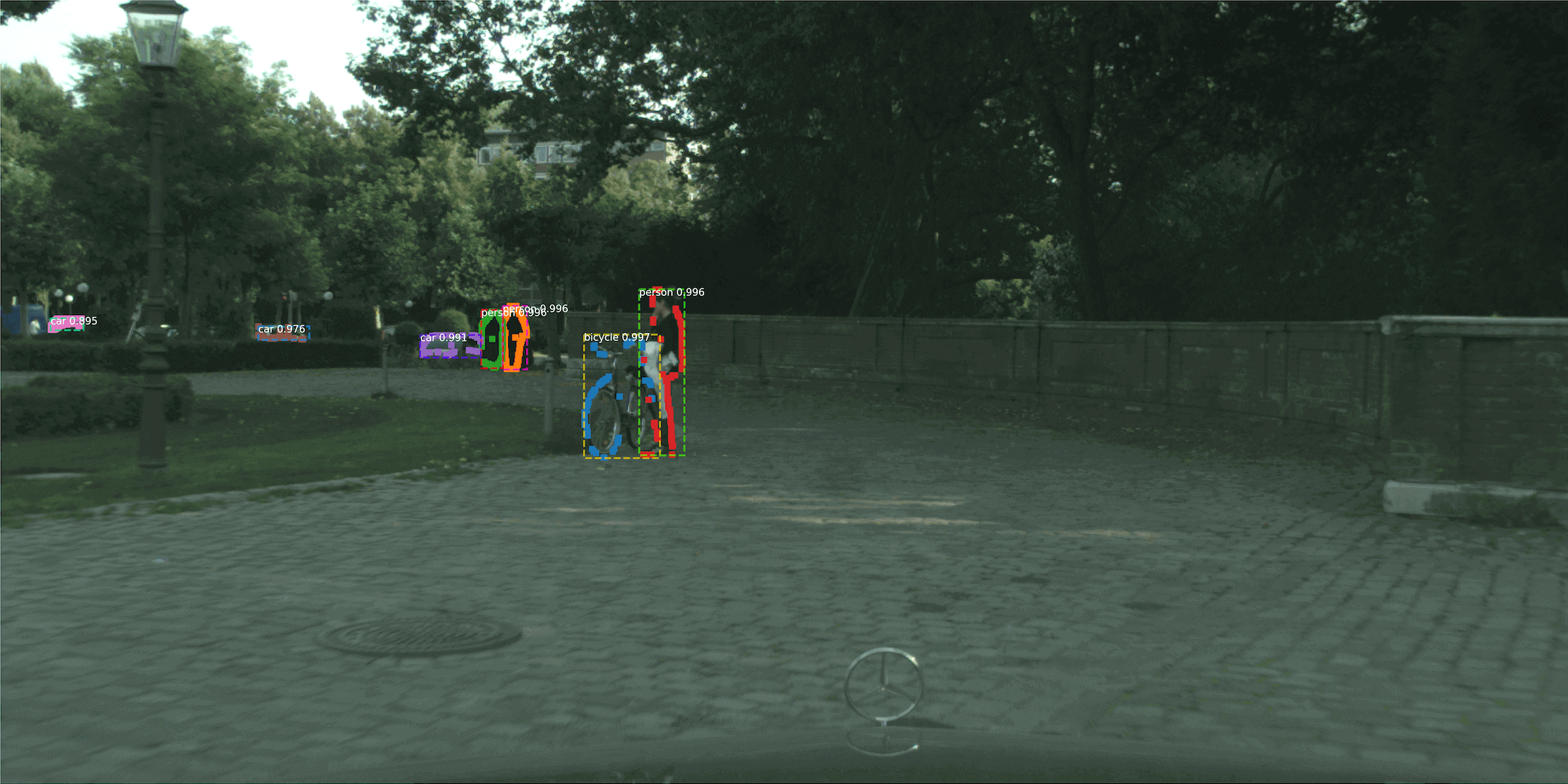}} \\
    \subfigure{\includegraphics[width=0.22\linewidth]{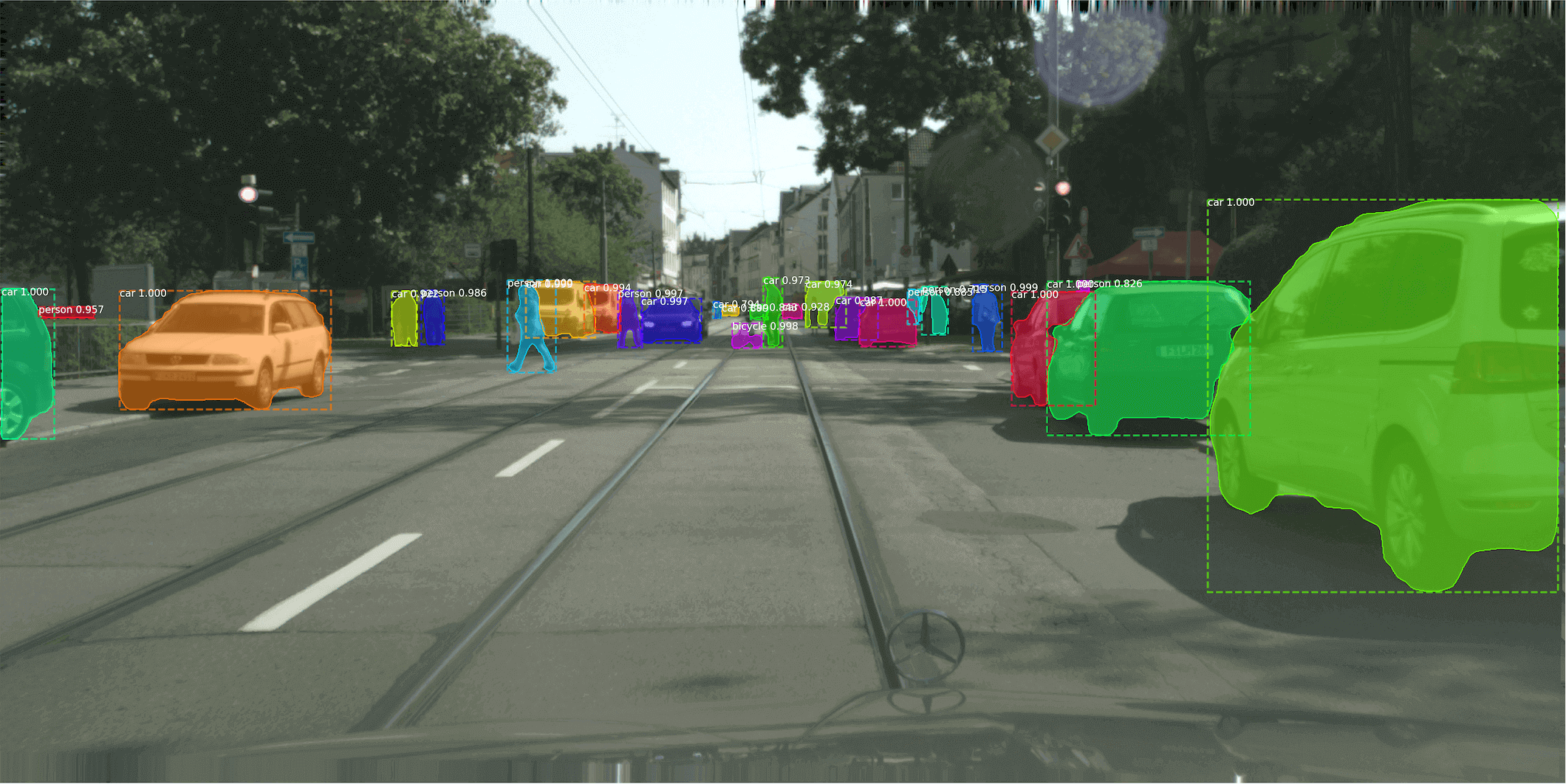}} 
    \subfigure{\includegraphics[width=0.22\linewidth]{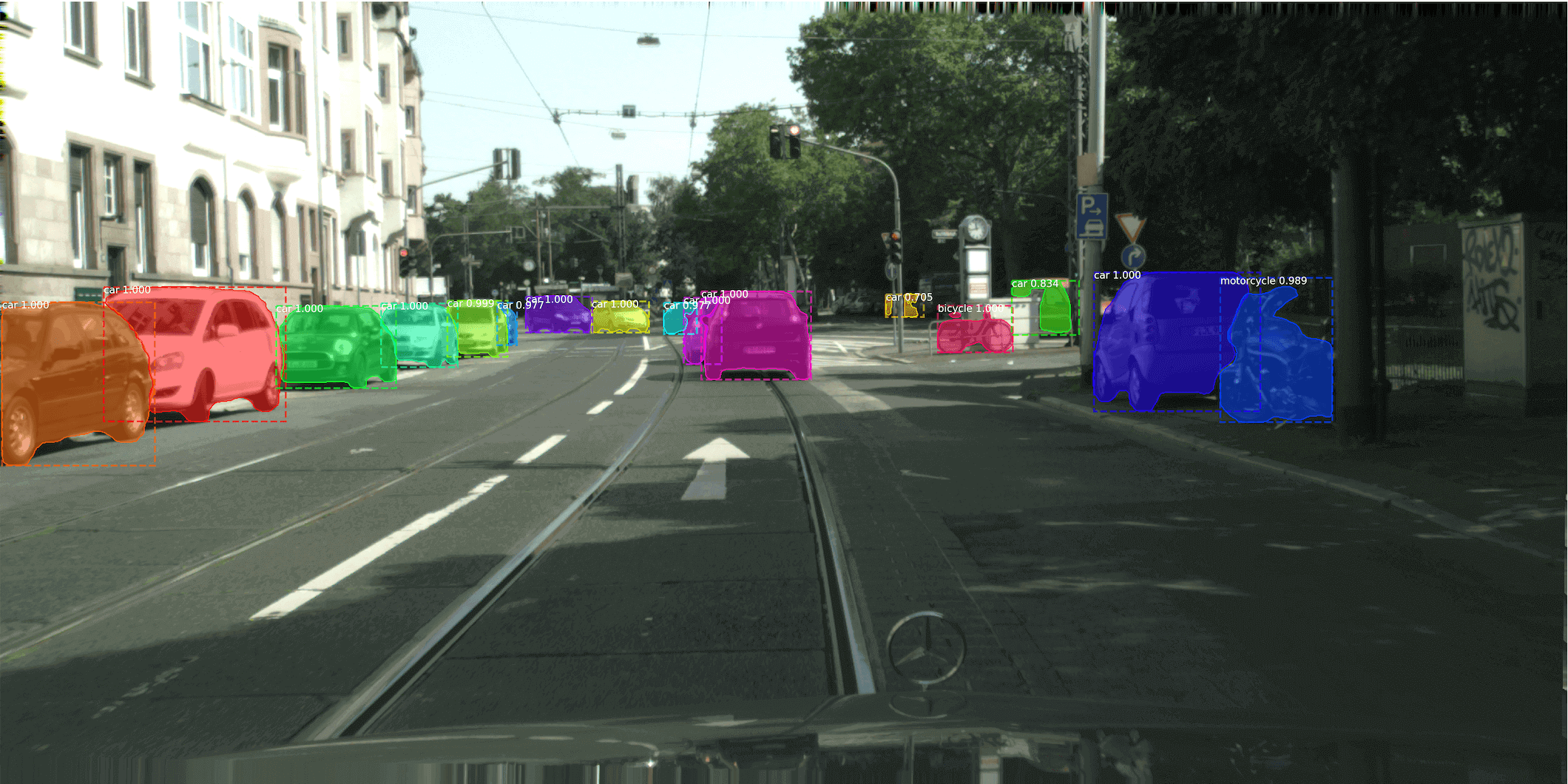}}  
  \subfigure{\includegraphics[width=0.22\linewidth]{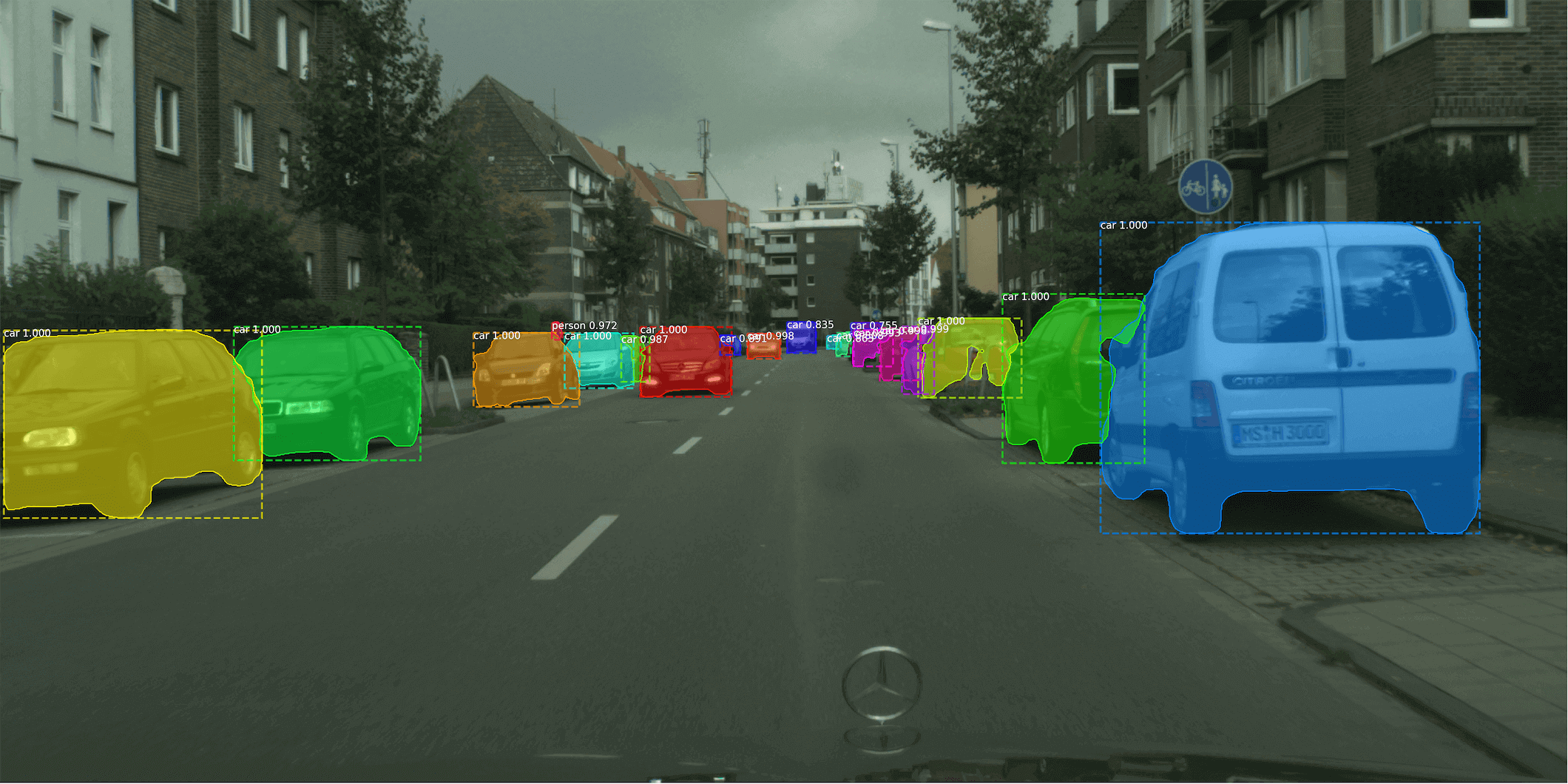}} 
    \subfigure{\includegraphics[width=0.22\linewidth]{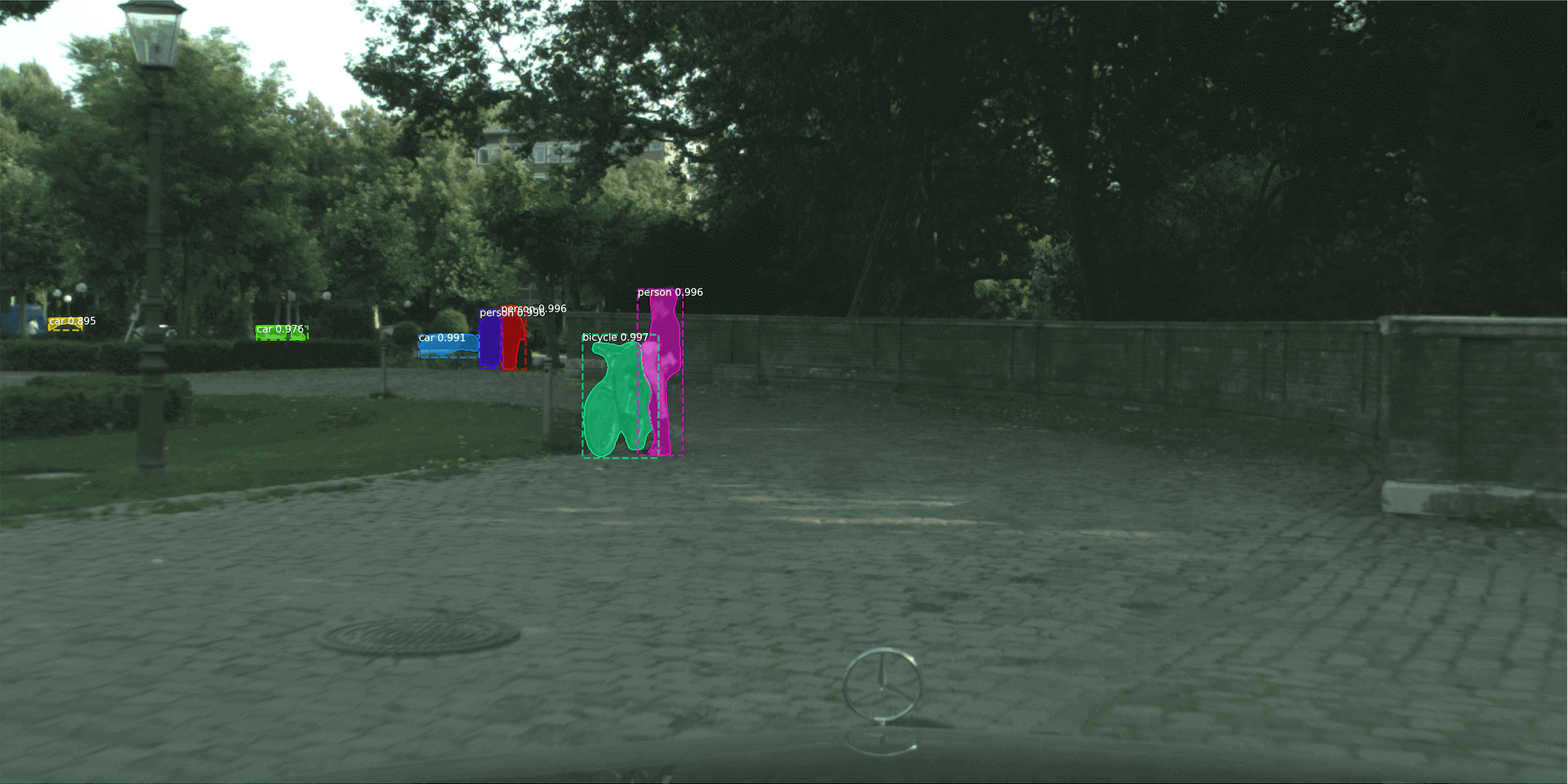}} 
    \caption{Instance segmentation results of our model on the Cityscapes validation set, top row: input images; middle row: keypoint detection results; bottom row: joint training detection results.}
  \label{fig:fig2}
\end{figure*}

\subsection{Motivation}
Since most of the existing instance segmentation methods are implemented by pixel-by-pixel classification, these methods depend on finer label information relative to the object detection task. For this reason, making a instance segmentation dataset will be a major expenditure of time and effort. During the annotation process, the annotator needs to obtain the edge contour of the instance by connecting the edge points for building the instance's mask, so the boundary point features of the instance can well represent its geometric information. We all know that deep learning is a typical data-driven method. Therefore, in order to improve the accuracy of instance segmentation when the amount of data is limited, inspired by the artificial annotating process, we obtain a fixed number of edge aggregation points by uniformly sampling the edge contours of the groundtruth mask, and then utilize the newly added keypoint prediction branch to predict these edge aggregation points. Finally, the binary segmentation mask is obtained by fusing the edge information prediction and the mask prediction. 

\subsection{Mask R-CNN}
 Next, we briefly review the framework of Mask R-CNN. Mask R-CNN is extended on the basis of Faster R-CNN, in addition to the shared backbone network and FPN (Feature Pyramid Networks), it in parallel adds a new branch for predicting the object mask on the bounding box recognition branch. The network can also be easily extended to other tasks, such as person keypoint detection (i.e. Keypoint R-CNN). In Keypoint R-CNN, its keypoint head is very similar to the mask head of Mask R-CNN. The mask head obtains an output resolution of $28\times28$ through four $3\times3$ convolutional layers with a channel of 256 which followed by an up-sampling layer, and finally employ a convolutional layer to make its dimensions consistent with the object category. Meanwhile, the keypoint head obtains an output resolution of $56\times56$ through eight $3\times3$ convolutional layers with 512 channels and followed by two upsampling layers. Besides, in order to implement Human Pose Estimation in Mask R-CNN, it uses one-hot mask to encode each keypoint position, and the structural design is similar to the mask branch, which uses fully convolutional network to predict a binary mask for each keypoint.

\subsection{Boundary point extraction}
\begin{figure}[!htb]
  \centering
    \subfigure[]{\includegraphics[width=0.3\linewidth]{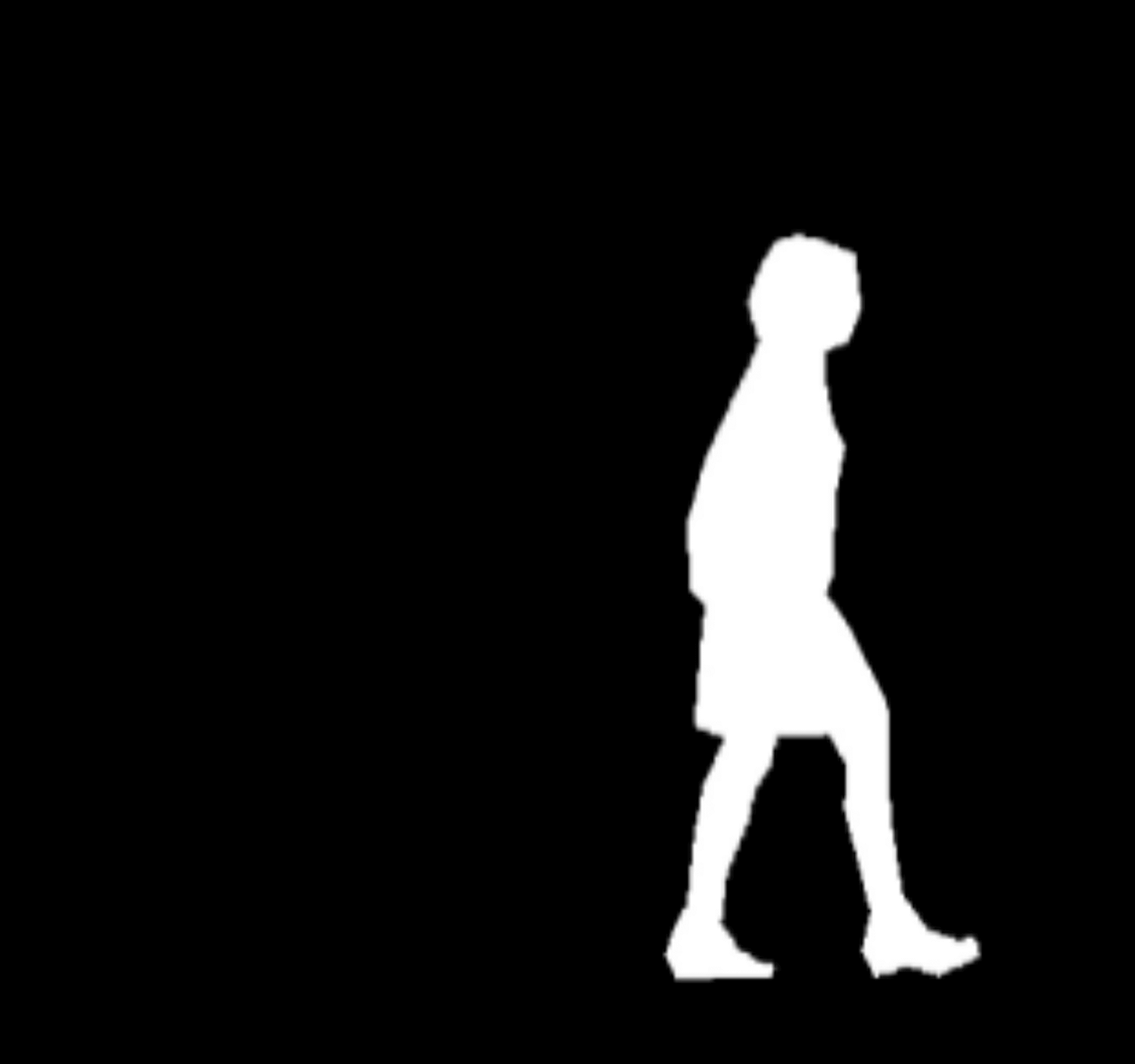}}  
  \subfigure[]{\includegraphics[width=0.3\linewidth]{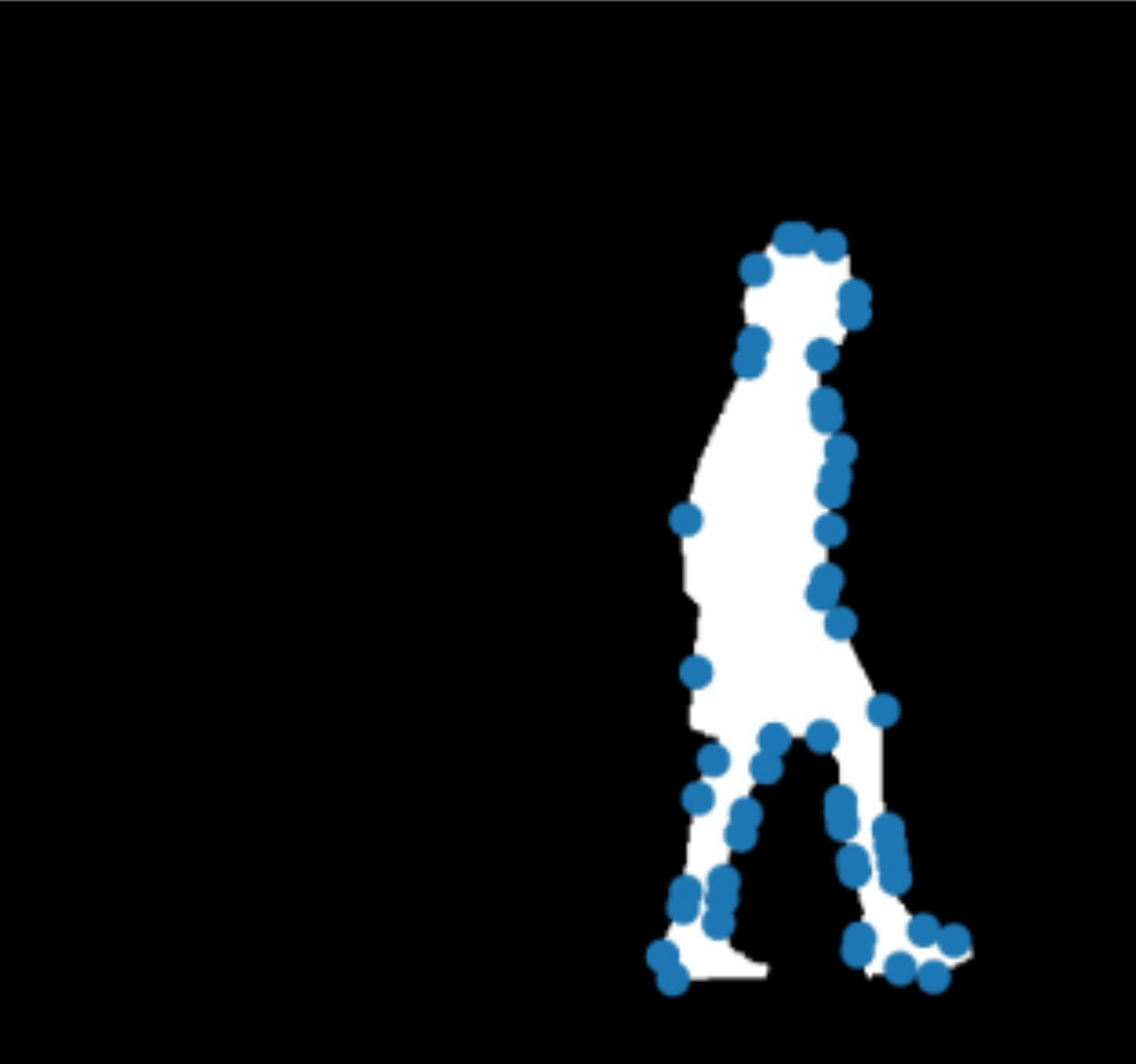}} 
    \subfigure[]{\includegraphics[width=0.3\linewidth]{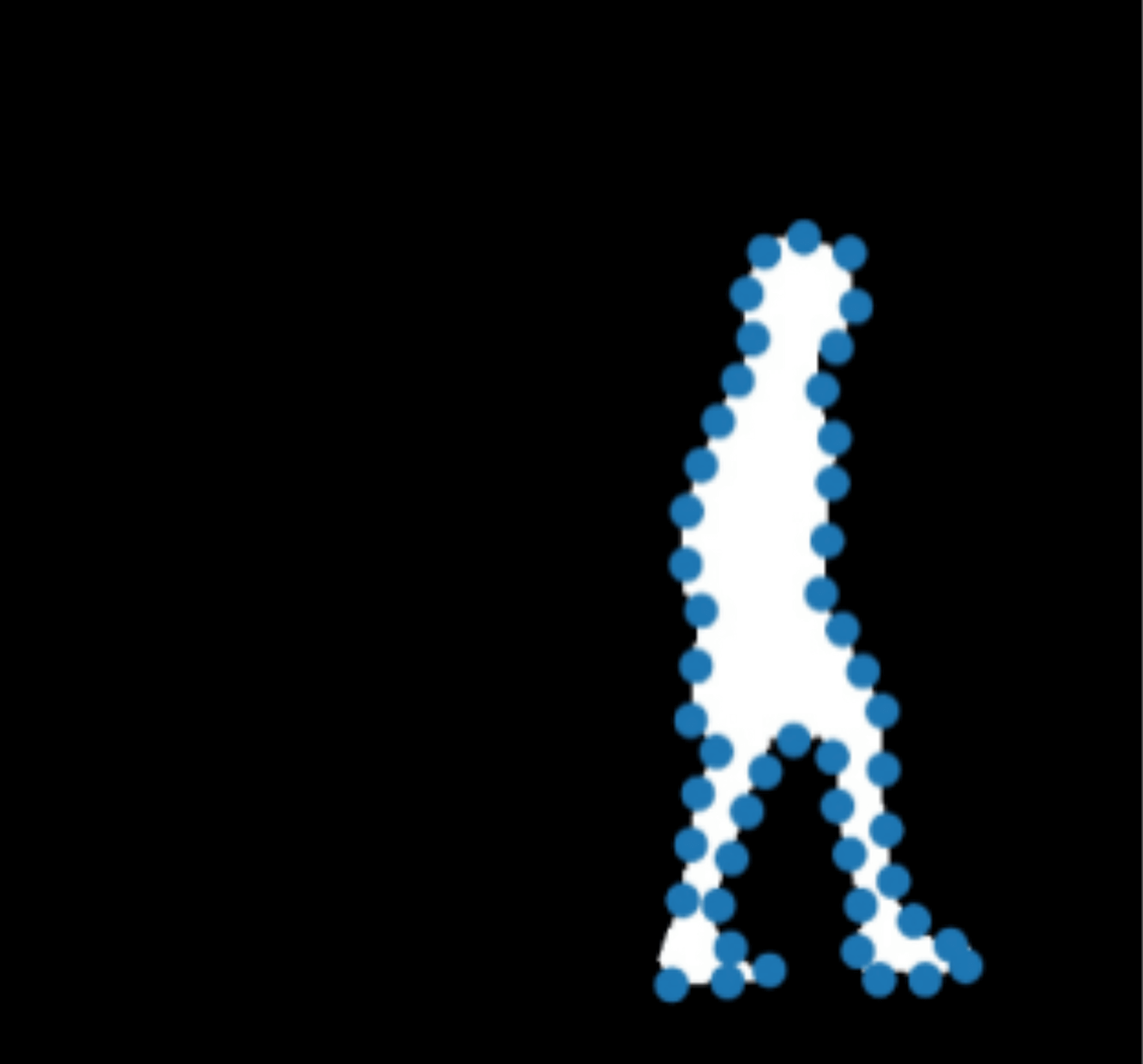}} 
    \caption{Examples of object contour point sampling. (a) Target binary mask label. (b) Corner sampling. (c) Uniform sampling.}
  \label{fig:fig3}
\end{figure}

To train our keypoint detection subtask, we need the labels of the object edge points. But the dataset used for instance segmentation only provides binary groundtruth mask, so we need to make labels for keypoint detection subtasks ourselves. In the keypoint detection task, a heat map is generated for each detection point. Thus, for different instance targets in the training data, we need to extract the contour boundary points of the object to form training labels. However, the CNN (Convolutional Neural Network) network require the input training data with a constant channel dimension, hence we need to extract a fixed number of boundary points correspondingly. In this work, we generate training labels for keypoint detection by sampling the instance contours, as shown in the Fig.~\ref{fig:fig3}. But due to the large difference in the scale of the instances in the training dataset, the number of points obtained by sampling may be small. In this regard, we complement it to the required amount by randomly selecting existed sampling points. Besides, we also explored the different sampling method, which will be explained in detail in the experimental section.

\begin{table*}[]
\begin{center}
\caption{ Results on Cityscapes val subset (denoted as AP {[}val{]}), and test subset (denoted as AP), * means our re-implementation}
\begin{tabular}{@{}cccccccccccc@{}}
\toprule
Methods & AP{[}val{]} & AP & AP$_{50}$ & person & rider & car & truck & bus & train & motorcycle & bicycle \\ \midrule
Kendall $et~al$.~\cite{kendall2018multi}              & -                               & 21.6                   & 39.0                     & 19.2                       & 21.4                      & 36.6                    & 18.8                      & 26.8                    & 15.9                      & 19.4                           & 14.5                        \\
Arnab $et~al$.~\cite{arnab2017pixelwise}                & -                               & 23.4                   & 45.2                     & 21.0                       & 18.4                      & 31.7                    & 22.8                      & 31.1                    & 31.0                      & 19.6                           & 11.7                        \\
SGN~\cite{liu2017sgn}                        & -                               & 25.0                   & 44.9                     & 21.7                       & 20.0                      & 39.4                    & 24.7                      & 33.2                    & 30.8                      & 17.7                           & 12.3                        \\
PolygonRNN++~\cite{acuna2018efficient}                & -                               & 25.4                   & 45.4                     & 29.3                       & 21.7                      & 48.2                    & 21.1                      & 32.3                    & 23.7                      & 13.6                           & 13.6                        \\
Neven $et~al$.~\cite{neven2019instance}                & -                               & 27.6                   & 50.8                     & 34.5                       & 26.0                      & 52.4                    & 21.6                      & 31.1                    & 16.3                      & 20.0                           & 18.8                        \\
GMIS~\cite{liu2018affinity}                        & -                               & 27.6                   & 44.6                     & 29.2                       & 24.0                      & 42.7                    & 25.3                      & \textbf{37.2}                    & \textbf{32.9}                      & 17.5                           & 11.8                        \\
BshapeNet+ {[}fine-only{]}~\cite{kang2020bshapenet}  & -                               & 27.3                   & 50.4                     & 29.7                       & 23.3                      & 46.7                    & 26.0                      & 33.3                    & 24.8                      & 20.3                           & 14.1                        \\
Mask R-CNN{[}fine-only{]}~\cite{he2017mask}   & 31.5                            & 26.2                   & 49.9                     & 30.5                       & 22.7                      & 46.9                    & 22.8                      & 32.2                    & 18.6                      & 19.1                           & 16.0                        \\
Mask R-CNN*{[}fine-only{]}~\cite{he2017mask}   & 33.1                            & 28.5                   & 55.0                     & 34.0                       & 25.7                      & 51.3                    & 24.4                      & 32.1                    & 22.1                      & 20.9                           & 17.5                        \\

Ours{[}fine-only{]}         & \textbf{36.9}                            & \textbf{31.2}                   & \textbf{56.1}                     & \textbf{37.3}                       & \textbf{28.2}                      & \textbf{55.9}                    & \textbf{28.6}                      & 35.1                    & 23.7                      & \textbf{21.3}                           & \textbf{19.8}  \\
 \bottomrule
\end{tabular}

\label{tab:my-table1}
\end{center}
\hspace{-5.0em}
\end{table*}

\subsection{Model framework}
According to the definition in~\cite{zhang2017survey}, we can use symbols $
\{\tau_{i}\}_{i=1}^m $ to describe multi-task training deep learning models, where $m$ represents the number of subtasks. For our framework $m=4$, it includes the bounding box regression task, the classification task, the target segmentation task, and the newly added keypoint detection task. Suppose that our dataset $D$ contains a total of $n_{i}$ training samples, then $D=\{x_{i}^{j}, y_{i}^{j}\}$, where $x_{i}^{j}$ is the $j-th$ training instance in $\tau_{i}$ and $y_{i}^{j}$ is its corresponding training label. In our approach, we used the same training data for different learning tasks with different training labels, i.e. $x_{i}=x_{l}, y_{i}\not=y_{l}, and (i, l)\in\mathbb{R}^{m}$. More formally, the training label for the regression task of the bounding box is the coordinate of the center point of the groundtruth and its width and height. And for the target segmentation task, the training label is the corresponding binary mask, whereas for the keypoint detection task, the training label is the edge aggregation point obtained by sampling the target boundary.

Fig.~\ref{fig:fig1} shows the general architecture of our model. In our pipeline, the box head and mask head are standard components of original Mask R-CNN. Obviously, we can simultaneously perform keypoint joint training with other heads by adding the keypoint detection branch. Suppose that we utilize $k$ edge aggregation points and the center point of the object, a total of $k+1$ points, as the keypoint labels information for training. And then a $M \times M$ heat map will be generated for each keypoint in the keypoint detection task of Mask R-CNN, so we can get a $M \times M \times(k+1)$ output mask tensor eventually. As for the mask prediction branch, a $N \times N \times C$ output tensor is produced through the fully convolutional network, where $C$ represents the total number of object categories. In addition, for capturing the overall edge information of the target, we do the channel-wise addition on the output features of the keypoint prediction branch for mapping the information of multiple keypoint with one hot encoding onto a single heat map. And in the next steps we reduce the heat map into a feature map with the same spatial scale as the mask prediction branch output by applying a $3\times3$ convolution layer. In the end, we broadcast the edge information of the target to each channel of the mask prediction branch through feature fusion operation. There is one more thing should notice that we do not use ReLU (Rectified Linear Unit) activation function after the $3\times3$ downsampling convolutional layer. As a consequence, our model can pay more attention to the segmentation edge of the object through the direct prediction of the object boundary point and the indirect fusion with the object segmentation feature output by the mask head. 

The above fusion process could be mathematically described as follows:
\begin{displaymath}
O_k=\sum_{i=1}^{i=k+1}Conv_{3\times3}^{s=2}(f_{k+1}),\tag{$1$}
\end{displaymath}
\begin{displaymath}
O_m=O_k\otimes f_m,\tag{$2$}
\end{displaymath}
where $\otimes$ represents channel-wise multiplication, $s$ stands for the stride of the convolution operation, $f_{k+1}$ stands for the output features of the keypoint prediction branch, and $f_m$ stands for the output features of the mask prediction branch.

\subsection{Loss function}
In original Mask R-CNN, the multi-task loss is defined as three parts: $L_{cls}$, $L_{box}$, and $L_{mask}$. For our model we calculate the loss in the keypoint detection task by using the cross entropy function, which is  the same as  defined in Mask R-CNN. After the keypoint detection is added to our model, the loss function can be updated as follows:\begin{displaymath}
L=L_{cls}+L_{box}+L_{mask}+{\alpha}L_{keypoint},\tag{$3$}
\end{displaymath}
where $\alpha$ represents the weight parameter to balance the loss of the keypoint detection task.

\section{Experiments}
\label{expe}
In this section, we will first introduce our experimental dataset, evaluation criteria, and model parameter settings. Then we conduct ablation experiments from two aspects, namely the edge feature fusion and influence of keypoint. Moreover, we report the results of instance segmentation only using the edge features of the keypoint.

\subsection{Dataset and metrics}
We verify our algorithm on the Cityscapes dataset, which contains a variety of stereo video sequences recorded in street scenes from 50 different cities. It provides 5000 images with fine annotations of a fixed resolution of $2048\times1024$, which are splitted into 2975, 500 and 1525 images for train/val/test, and the test dataset labels are not publicly released. We evaluate results based on AP (Average precision) and AP$_{50}$. We conducte plenty of ablation experiments on the validation set and provided results on the test dataset by uploading our results to the server. We compare our model with other state-of-the-art methods on Cityscapes test subset, and the results are shown Table ~\ref{tab:my-table1}. Training on “fine-only” data, our method outperforms Mask R-CNN by 2.7\% on test subset. Moreover, in order to verify that our method is not limited to a specific dataset, we perform experiment on COCO dataset to verify the generalization of our method. And in COCO dataset, it provides 115k images for training and 5k images for verification, and 20k images for testing.

\subsection{Implementation Details}
We perform our experiments based on the torchvision detection module with a pytorch backend~\cite{paszke2019pytorch}. We take 4 images in one image batch for training. The shorter edges of the images are randomly sampled from [800, 1024] for reducing overfitting. Each training is carried out on two nvidia titan xp GPUs. For instance segmentation task the mask spatial scale is $28\times28$, and for keypoint detection task is $56\times56$. We randomly horizontal flip the input image with a probability of 0.5 to augment the dataset. Because we use the different number of GPUs, different training strategies are used for Citysacpes dataset compared with Mask R-CNN. We train our model a total of 64 epochs, with a learning rate of 0.005 for the fore 48 epochs and 0.0005 for the remaining 16 epochs. Following Mask R-CNN, we use SGD for gradient optimization with weight decay 0.0001 and momentum 0.9. ResNet-50 FPN~\cite{he2016deep} is taken as the initial model on this dataset, if not specially noted. And for keypoint detection task we use uniform sampling with 100 sampling points. For the COCO dataset we train our model for 24 epochs, with an initial learning rate of 0.005, and reduce it by 0.1 after 16 and 22 epochs, respectively.

\begin{figure*}[!htb]
  \centering
    \subfigure[]{\includegraphics[width=0.35\linewidth]{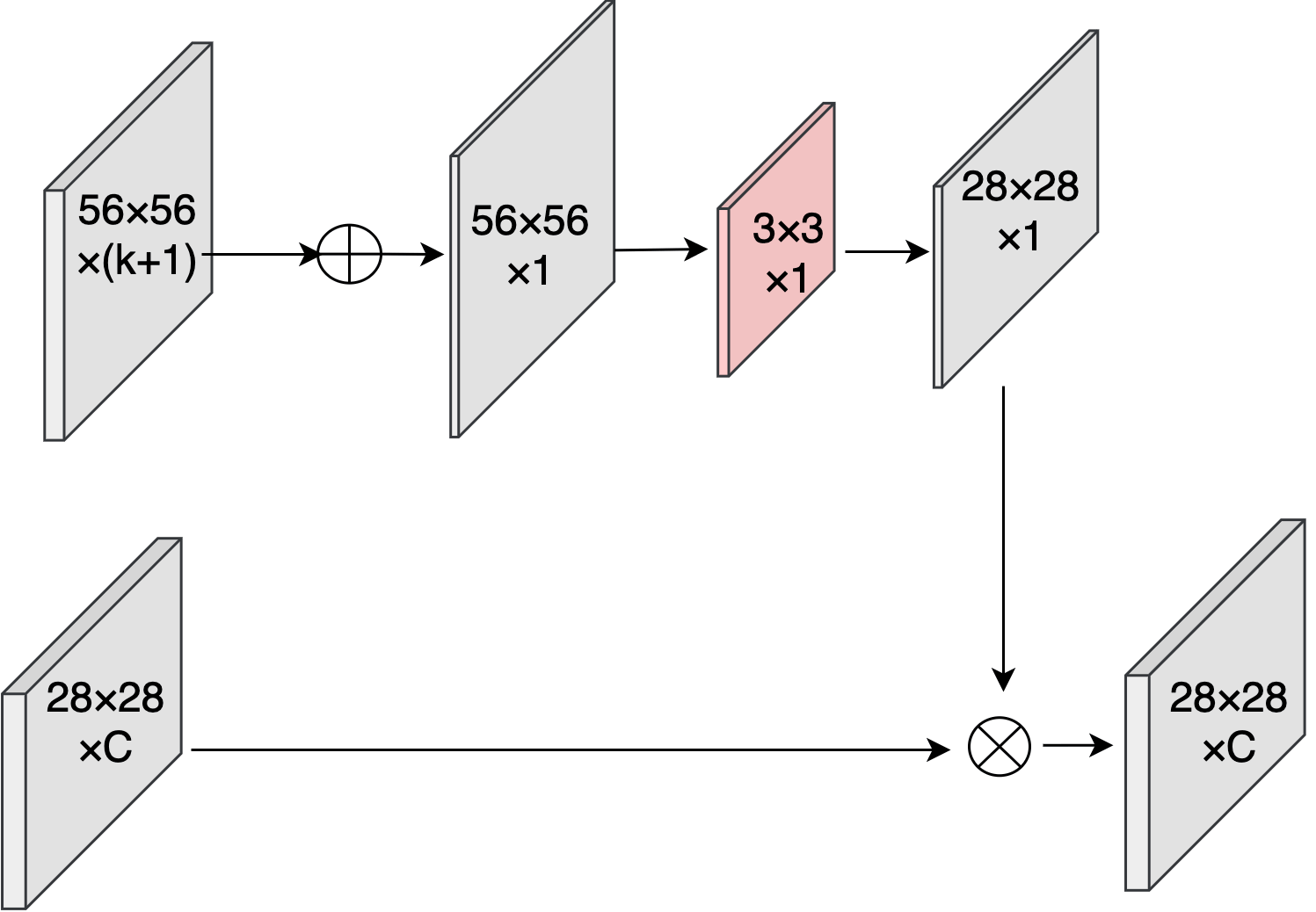}}  \quad \quad
  \subfigure[]{\includegraphics[width=0.35\linewidth]{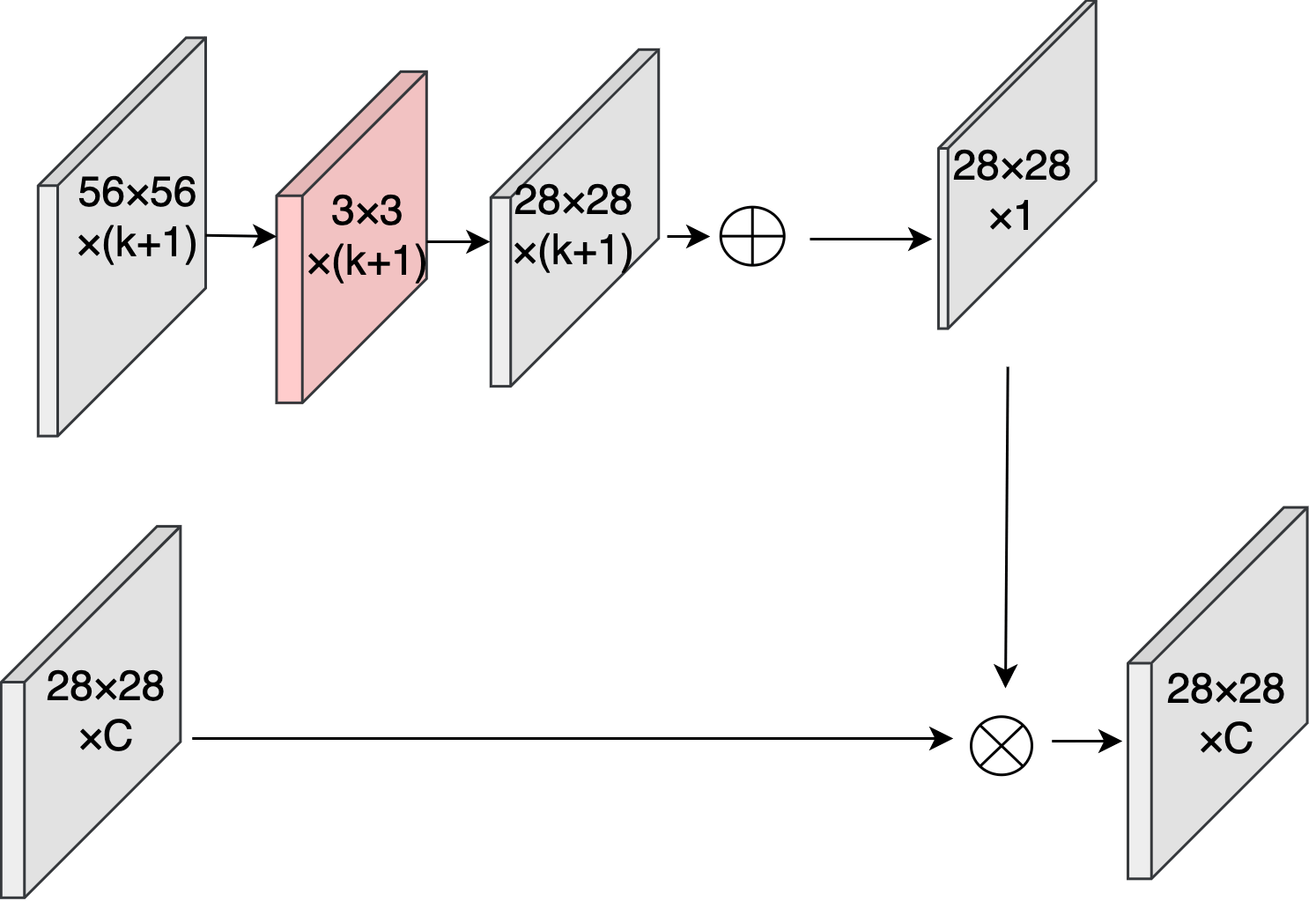}} \\
    \subfigure[]{\includegraphics[width=0.25\linewidth]{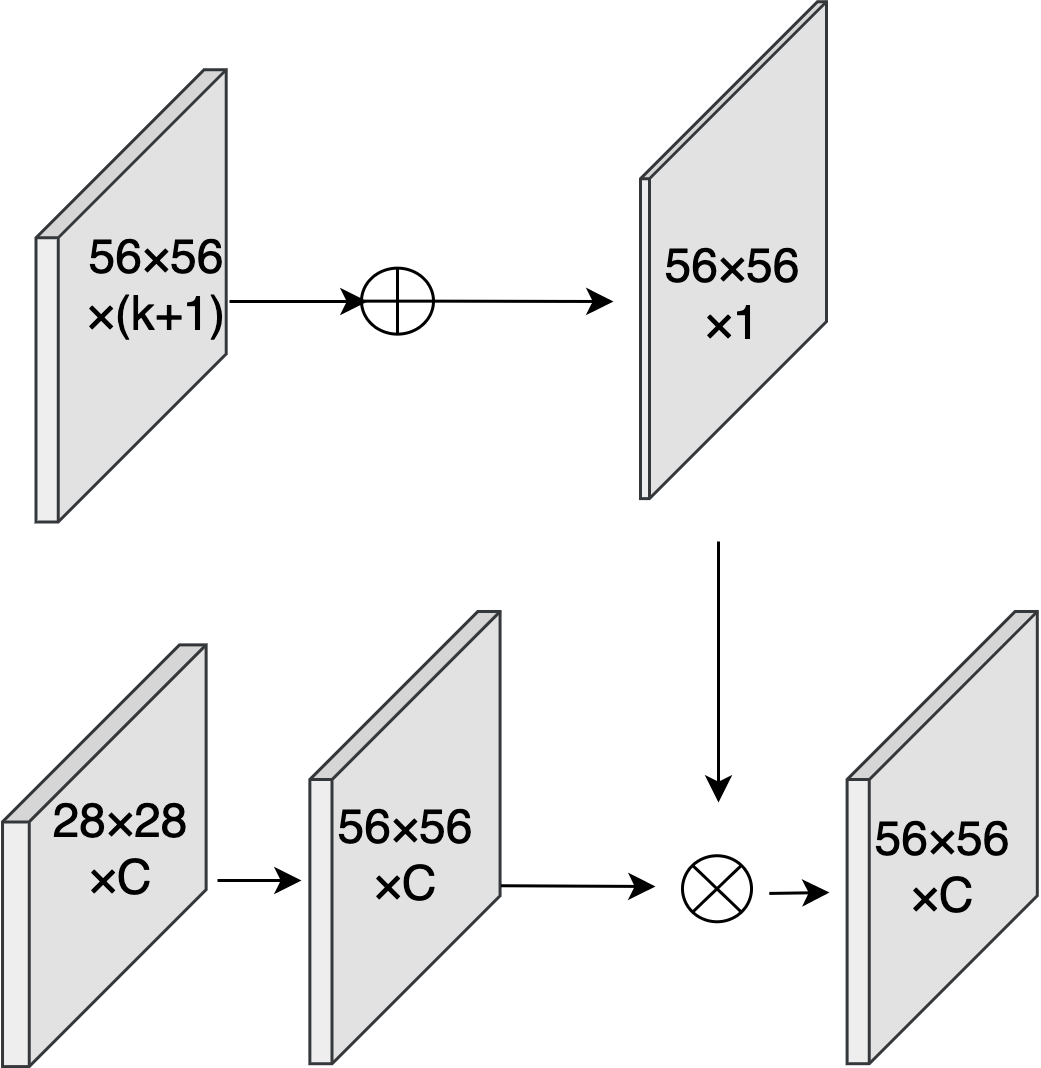}} \quad \quad \quad \quad
  \subfigure[]{\includegraphics[width=0.38\linewidth]{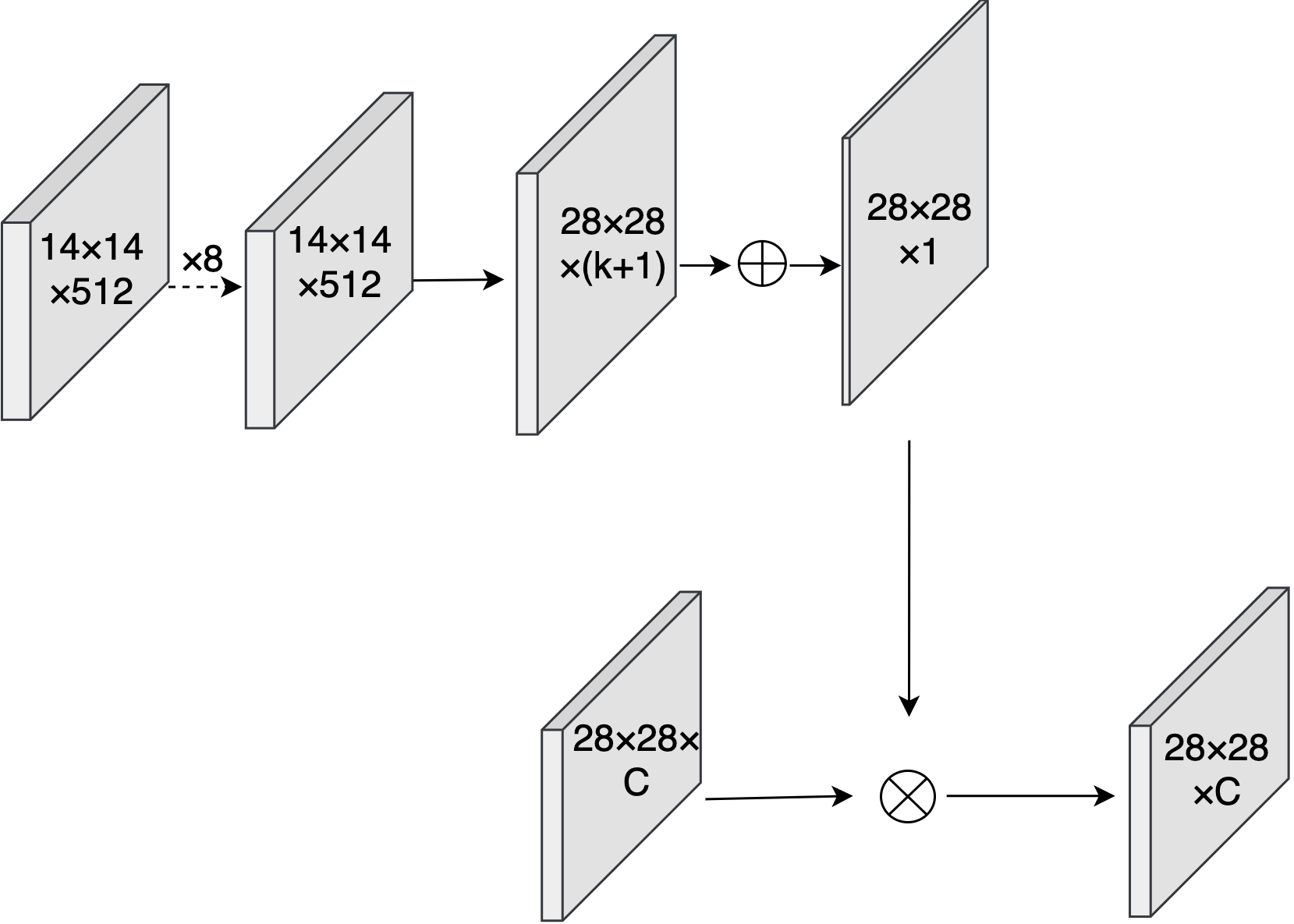}} \\
  \caption{Different structural choices for feature fusion.}
  \label{fig:fig4}
\end{figure*}

\subsection{Edge Feature Fusion}
{\bf The structural design choices for feature fusion:}
In our framework, feature fusion between multiple tasks is required, but the mask spatial scale of instance segmentation task and keypoint detection task is different, $56\times56$ vs $28\times28$. Therefor, we need to ensure that the input features of the fusion operation have the same spatial scale. There are a few design choices shown in Fig.~\ref{fig:fig4} and explained as follows:

\begin{table}[]
\centering
\caption{Results of different structural designs on Cityscapes val subset}
\setlength{\tabcolsep}{4mm}{
\begin{tabular}{ccc}
\toprule
Setting    & AP   & AP$_{50}$ \\ \midrule
Mask R-CNN re-implement & 33.1 & 60.0 \\
   design (a)        & 31.8 & 59.5 \\
   design (b)        & \textbf{36.0} & \textbf{63.0} \\
   design (c)        & 35.6 & 62.3 \\
   design (d)        & 34.9 & 62.1 \\ \bottomrule
\end{tabular}}
\label{tab:my-table2}
\vspace{-1.5em}
\end{table}

\begin{table}[]
\centering
\caption{Results of different spatial scale reduction desiges on Cityscapes val subset}
\setlength{\tabcolsep}{4mm}{
\begin{tabular}{ccc}
\toprule
Setting        & AP   & AP$_{50}$ \\ \midrule
Mask R-CNN re-implement    & 33.1 & 60.0 \\
maxpooling     & 34.1 & 61.5 \\
avgpooling     & 33.6 & 60.7 \\
3x3 convlution & \textbf{36.0} & \textbf{63.0} \\ \bottomrule
\end{tabular}}
\label{tab:my-table3}
\vspace{-1.5em}
\end{table}

\begin{table}[]
\centering
\caption{Results of different fusion strategies on Cityscapes val subset}
\setlength{\tabcolsep}{4mm}{
\begin{tabular}{ccc}
\toprule
Setting    & AP   & AP$_{50}$ \\ \midrule
Mask R-CNN re-implement& 33.1 & 60.0 \\
add        & 34.3 & 61.5 \\
max        & 31.9 & 61.0 \\
multiply   & \textbf{36.0} & \textbf{63.0} \\ \bottomrule
\end{tabular}}
\label{tab:my-table4}
\vspace{-1.5em}
\end{table}

\begin{table}[]
\centering
\caption{Results of different contour sampling points on Cityscapes val subset}
\setlength{\tabcolsep}{2mm}{
\begin{tabular}{ccc}
\toprule
Number of samples & AP   & AP$_{50}$ \\ \midrule
Mask R-CNN re-implement        & 33.1 & 60.0 \\
k = 50            & 34.5 & 61.7 \\
k = 100           & \textbf{36.0} & \textbf{63.0} \\ 
k = 150           & 35.3 & 62.1 \\ \bottomrule
\end{tabular}}
\label{tab:my-table5}
\end{table}

\begin{table}[]
\centering
\caption{Results of different contour sampling methods on Cityscapes val subset}
\setlength{\tabcolsep}{4mm}{
\begin{tabular}{ccc}
\toprule
Type of samples & AP   & AP$_{50}$ \\ \midrule
Mask R-CNN re-implement      & 33.1 & 60.0 \\
corner          & 35.1 & 62.1 \\
uniform         & \textbf{36.0} & \textbf{63.0}  \\ \bottomrule
\end{tabular}}

\label{tab:my-table6}
\vspace{-1.5em}
\end{table}

\begin{itemize}
\item[](a)\quad The output of keypoint predictor is size of $56 \times 56 \times (k + 1)$, and the output of mask predictor is size of $28 \times 28 \times C$. After we do channel-wise addition on the keypoint predictor output, the feature map spatial scale is reduced to $28 \times 28$ by using a $3 \times 3$ convolution layer with a stride of 2.

\item[](b)\quad The output of keypoint predictor and mask predictor is the same as that of design (a). The difference is before we do channel-wise addition on the keypoint predictor output, the feature map spatial scale is reduced to $28 \times 28$ by using a $3 \times 3$ convolution layer with a stride of 2.

\item[](c)\quad The output of keypoint predictor is size of $56 \times 56 \times (k + 1)$, and the output of mask predictor is size of $28 \times 28 \times C$. we upsample the output spatial scale of the mask predictor to $56 \times 56$ using transposed convolutional layer.

\item[](d)\quad The output of keypoint predictor is size of $28 \times 28 \times (k + 1)$, and the output of mask predictor is size of $28 \times 28 \times C$.
\end{itemize}

The experimental results are shown in Table~\ref{tab:my-table2}. We can see that each methods can achieve performance improvement except method (a), because in method (a) we only use one convolution layer to process the prediction results with a single channel, which will lose a lot of original edge prediction information.

{\bf The spatial scale reduction designs for feature fusion: }
Based on the design of the edge fusion structure, we try to reduce the spatial scale of the keypoint prediction output by using three methods: max pooling, average pooling, and a convolution layer with kernel size of 3 and stride of 2.

Table~\ref{tab:my-table3} shows the results for above spatial scale reduction designs, we can see that all the three methods can improve the performance of the model, which further validate the effectiveness of our method. Notably, we can see that compared with a simple pooling operation, for our model, using a convolutional downsampling method with learnable parameters performs better.

{\bf The choices of the fusion mode: }
After channel-wise addition and spatial scale reduction of the keypoint branch output, we can perform the fusion operation. In our experiment, we test three methods of maximum, multiplication, and addition to choose the best fusion strategy. Results are shown in Table~\ref{tab:my-table4}, we can see that the fusion method of multiplication can be better adapted to our model.

\begin{figure*}[!htb]
  \centering
    \subfigbottomskip=0.1pt
    \subfigure{\includegraphics[width=0.10\linewidth]{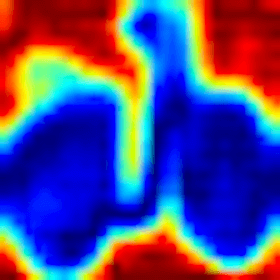}}  
  \subfigure{\includegraphics[width=0.10\linewidth]{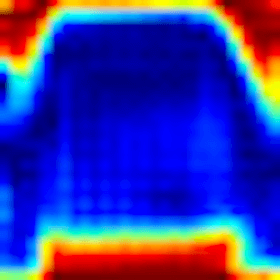}} 
  \subfigure{\includegraphics[width=0.10\linewidth]{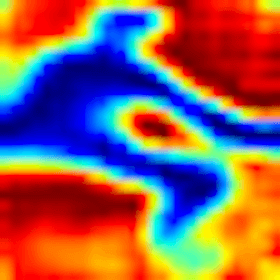}}
  \subfigure{\includegraphics[width=0.10\linewidth]{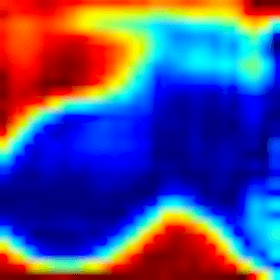}}  
  \subfigure{\includegraphics[width=0.10\linewidth]{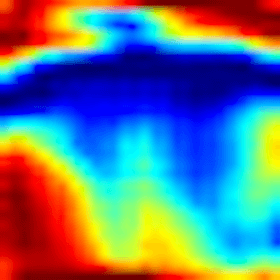}} 
    \subfigure{\includegraphics[width=0.10\linewidth]{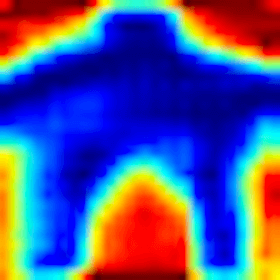}} \\
    \subfigure{\includegraphics[width=0.10\linewidth]{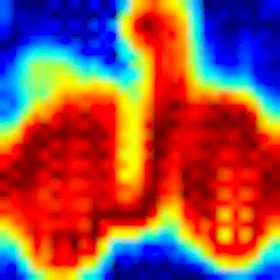}} 
  \subfigure{\includegraphics[width=0.10\linewidth]{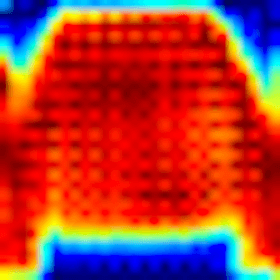}}
  \subfigure{\includegraphics[width=0.10\linewidth]{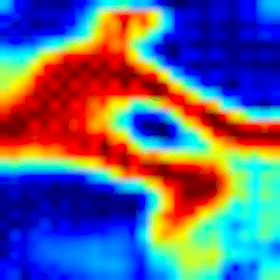}}
  \subfigure{\includegraphics[width=0.10\linewidth]{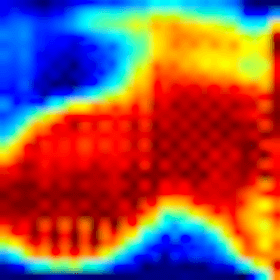}}
  \subfigure{\includegraphics[width=0.10\linewidth]{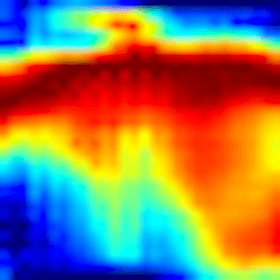}}   
    \subfigure{\includegraphics[width=0.10\linewidth]{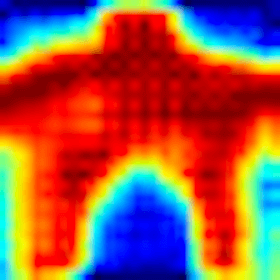}} 
    \caption{Examples of instance heat map, top row: feature maps output by keypoint head; bottom row: feature maps output by mask head.}
  \label{fig:fig5}
\end{figure*}

\subsection{Influence of keypoint}
The model we designed requires training tasks including box head, mask head, and keypoint head simultaneously. But the Cityscapes dataset only provides instance groundtruths masks, so we need to get the bounding box and edge aggregation point labels of the object from the mask of the instance ourselves. For a single target, the bounding box label is uniquely fixed, but for the edge aggregation point label we need to sample it from the edge contour of the target, which means that we have more alternative strategies. Specifically, we separately analyze the impact of corner sampling and uniform sampling on our method, and the results are shown Table~\ref{tab:my-table6}. Analogously, corner sampling reflects the key geometric information of the object, whereas uniform sampling reflects the overall geometric information of the object. From the experimental results we can see that it is more conducive to the instance segmentation by using keypoint detection technology to capture the overall geometric position information of the object. Additionally, for the case of the number of edge sampling points is less than the required number due to the target being too small when sampling the edge points, we randomly select the existing sampling points for padding.

\begin{table}[]
\centering
\caption{ Multi-task learning of box, mask, and keypoint, evaluated on Cityscapes val subset by using COCO evaluation metrics}
\setlength{\tabcolsep}{2mm}{
\begin{tabular}{cccc}
\toprule
Methods                   & AP$^{bb}$ & AP$^{mask}$ & AP$^{keypoint}$ \\ \midrule
Mask R-CNN, mask-only     & 37.7 & 32.4   & -    \\
Mask R-CNN, keypoint-only & 36.8    & -      & 55.7    \\
Ours (without center point) & 36.7 & 34.3   & 56.0 \\
Ours                       & 37.2 & 35.2   & 56.4 \\
Ours ($\alpha=0.5$) & \textbf{37.9} & \textbf{35.9}   & \textbf{58.2} \\ \bottomrule
\end{tabular}}

\label{tab:my-table7}
\end{table}

\begin{table}[]
\centering
\caption{Results of different balance factors on Cityscapes val subset}
\setlength{\tabcolsep}{12mm}{
\begin{tabular}{ccc}
\toprule
$\alpha$    & AP   & AP$_{50}$ \\ \midrule
1  & 36.0 & 63.0 \\
0.5 & \textbf{36.9} & \textbf{64.3}\\
0.2  & 35.1 & 62.1  \\ \bottomrule
\end{tabular}}

\label{tab:my-table8}
\end{table}

\begin{table*}[]
\caption{Comparisons with Mask R-CNN baseline on COCO datasets with ResNet-50 backbone}
\begin{center}
\setlength{\tabcolsep}{4mm}{
\begin{tabular}{ccccccccc}
\toprule
Methods     & AP$^{bb}$ & AP$^{bb}_{s}$ & AP$^{bb}_{m}$ & AP$^{bb}_{l}$ & AP$^{mask}$ & AP$^{mask}_{s}$ & AP$^{mask}_{m}$ &AP$^{mask}_{l}$ \\ \midrule
Mask R-CNN    & 37.9 & \textbf{21.7}   &41.3 &49.6 &34.6 &\textbf{15.8} &37.2 &51.1    \\
Ours  & \textbf{38.3} & 21.3 & \textbf{41.8}  &\textbf{50.3} &\textbf{35.4} &15.7 &\textbf{38.4} &\textbf{51.9}\\ \bottomrule
\end{tabular}}
\label{tab:my-table9}
\end{center}
\end{table*}

\begin{table*}[]
\caption{Comparisons with state-of-the-art counterparts on COCO test-dev dataset with ResNet-50 backbone}
\begin{center}
\setlength{\tabcolsep}{4mm}{
\begin{tabular}{cccccc}
\toprule
Methods    &Backbone & AP$^{mask}$ & AP$^{mask}_{s}$ & AP$^{mask}_{m}$ &AP$^{mask}_{l}$ \\ \midrule
Mask R-CNN~\cite{he2017mask}   &ResNet-50-FPN  & 34.9 & \textbf{19.0}   &37.4 &45.0 \\
Mask Scoring R-CNN~\cite{huang2019mask} &ResNet-50-FPN  & \textbf{35.8} & 16.2   &37.4 &\textbf{51.0}\\
YOLACT~\cite{bolya2019yolact} &ResNet-101-FPN & 31.2 & 12.1 & 33.3  &47.1 \\
CondInst~\cite{tian2020conditional} &ResNet-50-FPN & 35.4 & 18.4 & 37.9  &46.9 \\ 
BlendMask~\cite{chen2020blendmask} &ResNet-50-FPN & 34.3 & 14.9 & 36.4  &48.9 \\ 
PolarMask~\cite{xie2020polarmask} &ResNet-101-FPN & 32.1 & 14.7 & 33.8  &45.3\\  
Ours &ResNet-50-FPN & 35.7 & 18.6 & \textbf{38.1}  &46.4\\ \bottomrule
\end{tabular}}
\label{tab:my-table10}
\end{center}
\end{table*}

Furthermore, we demonstrate the influence of different sampling points number on the model's performance, the results are shown in Table~\ref{tab:my-table5}. Specifically, we conduct experiments on 50 and 100 sampling points respectively. And we find in the course of the experiment that the further increase of the sampling points number will damage the effectiveness of the model, which can be attributed to we add too much random information in padding operation, since there are a large number of small objects in the dataset.
Moreover, in our approach, we use not only the points sampled from the object contour but also the center point of the object when we perform the keypoint estimation task, which is follow the intuition that the geometric relationship between the object boundary point and the center point can be learned during the model training process for improving the estimation accuracy of the boundary point. For example, the center point of the object can be regarded as the origin in a polar coordinate system, for the objects with the same category, the distance and angle of the target boundary point relative to the center point have certain regularity. To verify our analysis, we evaluate the effect of adding the object center point using COCO evaluation metrics, as shown in the Table~\ref{tab:my-table7}, where AP$^{bb}$ represents the AP value of the bounding box detection results, AP$^{mask}$ and AP$^{keypoint}$ represent the AP value of the mask predict results and keypoint estimation results respectively, and the subscripts $s$, $m$, and $l$ represent small, medium, and large objects, respectively. We can see that our method can simultaneously improve the performance of instance segmentation, keypoint estimation and object detection tasks. 

\subsection{Influence of of the balance coefficient $\alpha$ of the loss function}
In our model, keypoint detection is only an auxiliary task, and we hope that it can complete its duties while not affecting other tasks as much as possible. Therefore, we balance the importance of each task in the multi-task joint training by setting the attenuation coefficient of its loss function. The experimental results are shown in the Table~\ref{tab:my-table8}, from the results in the Table, we can see that our method is very sensitive to the set balance coefficient, and we can get the best performance when we set $\alpha$=0.5.

\begin{figure*}[!htb]
  \centering
  \hspace{1em}\subfigure{\includegraphics[width=0.22\textwidth]{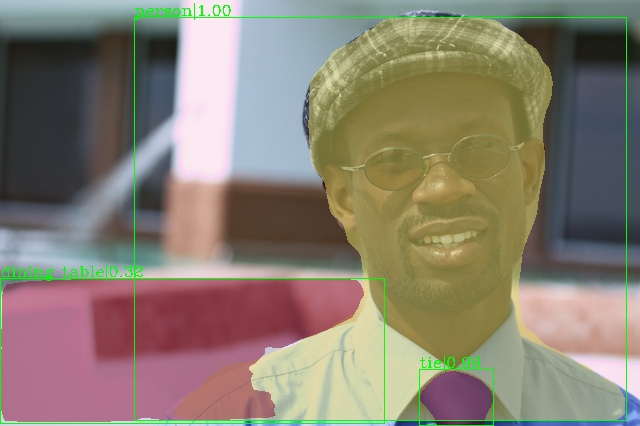}} \hspace{0.5em}
  \subfigure{\includegraphics[width=0.22\textwidth]{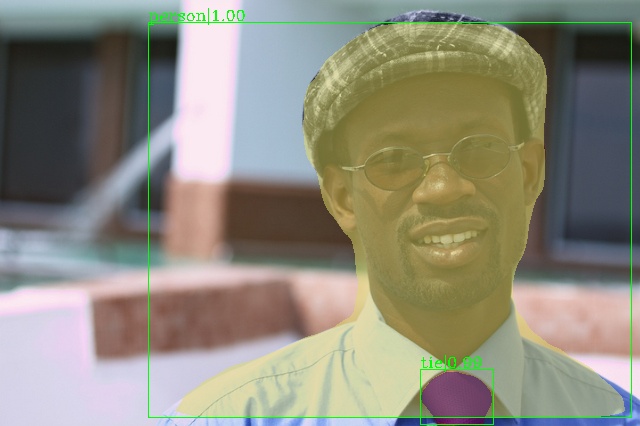}} \hspace{0.5em}
  \subfigure{\includegraphics[width=0.22\textwidth]{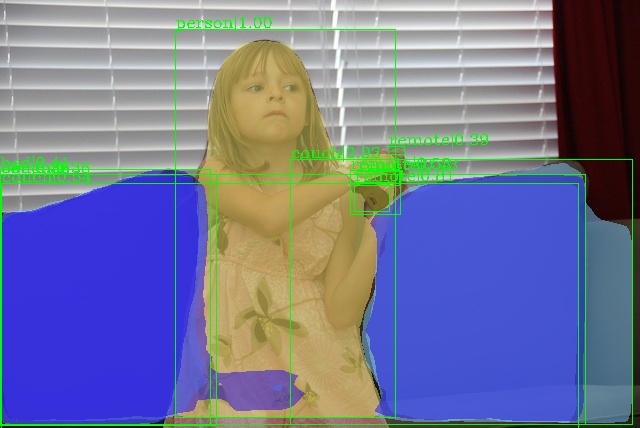}} \hspace{0.5em}
  \subfigure{\includegraphics[width=0.22\textwidth]{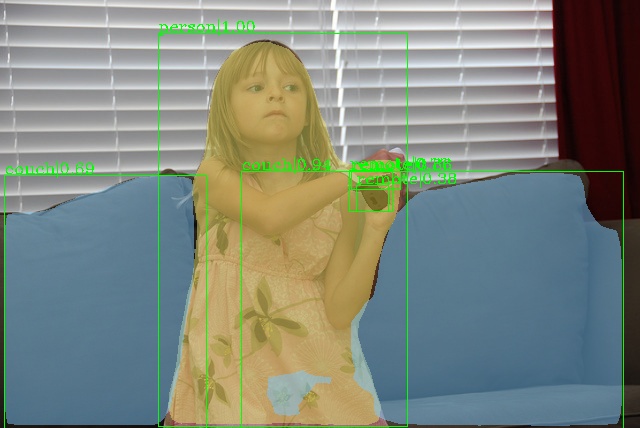}} \\
  \subfigure{\includegraphics[width=0.22\textwidth]{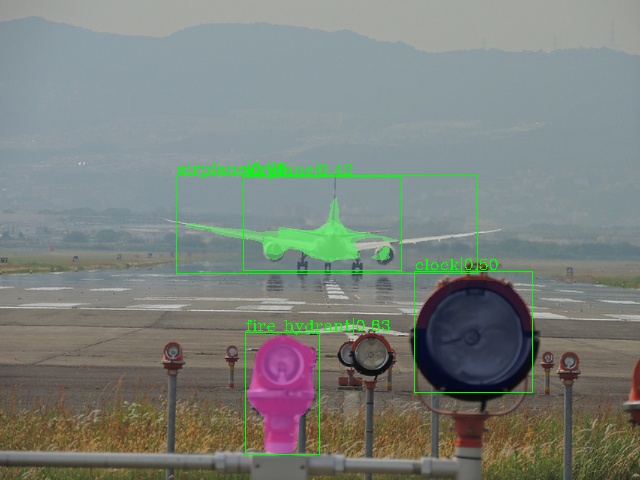}}  
  \subfigure{\includegraphics[width=0.22\textwidth]{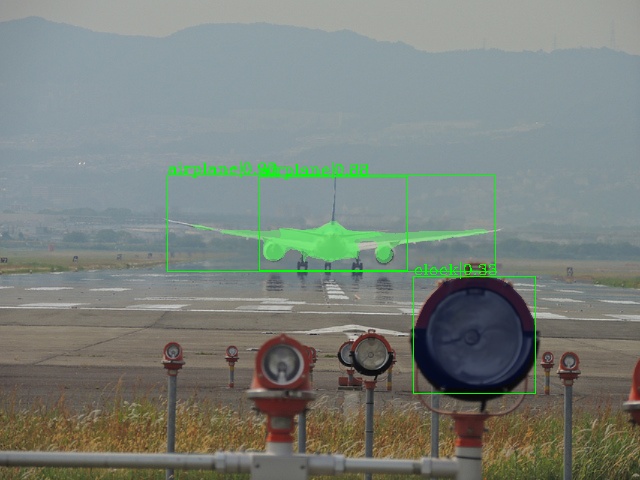}} 
  \subfigure{\includegraphics[width=0.24\textwidth]{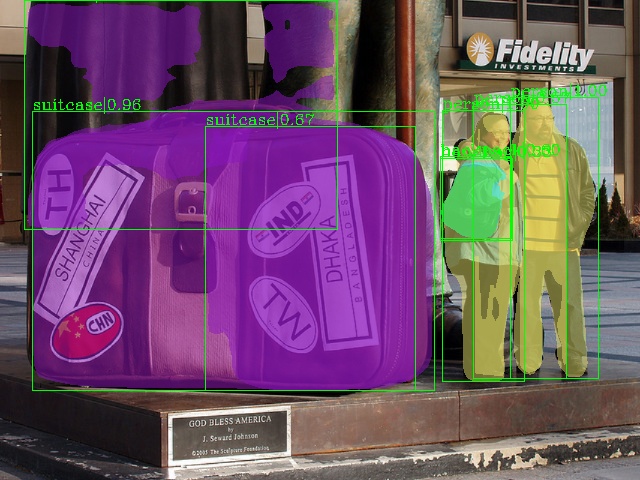}}  
  \subfigure{\includegraphics[width=0.24\textwidth]{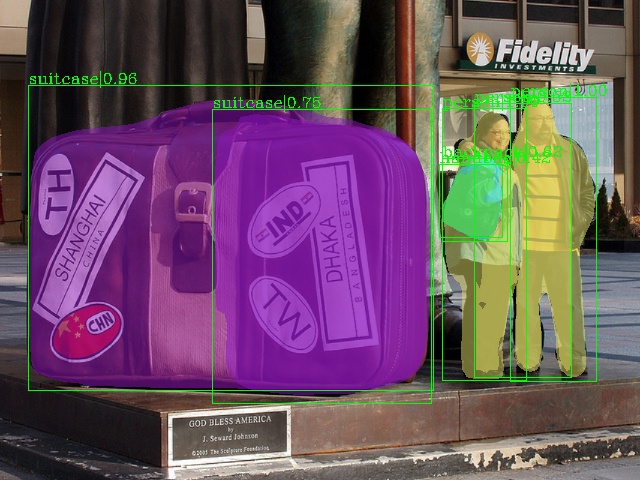}} \\
  \subfigure[Mask R-CNN Baseline]{\includegraphics[width=0.23\textwidth]{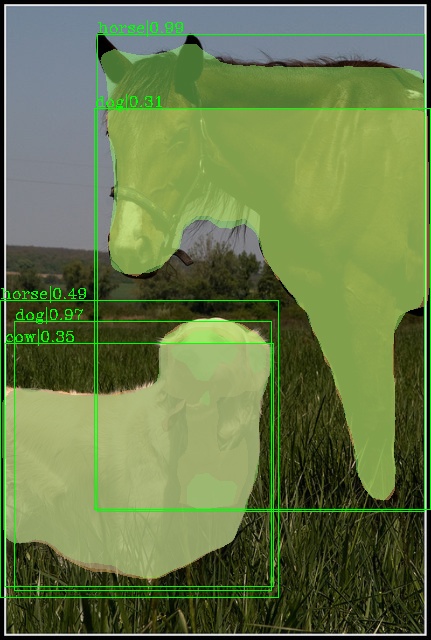}}  
  \subfigure[Ours]{\includegraphics[width=0.23\textwidth]{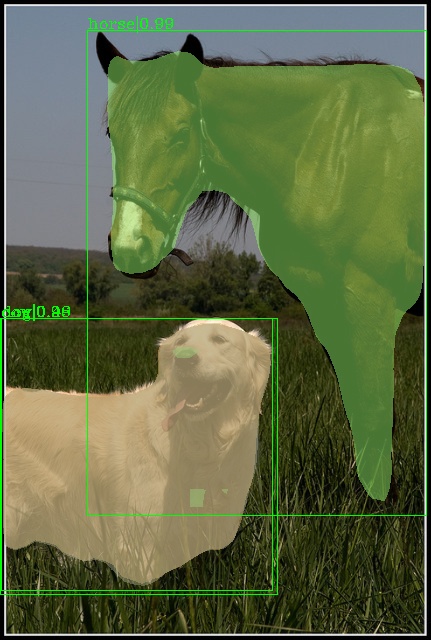}} 
  \subfigure[Mask R-CNN Baseline]{\includegraphics[width=0.23\textwidth]{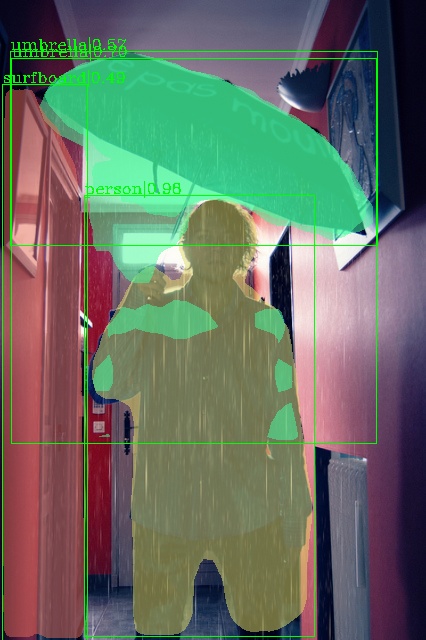}}  
  \subfigure[Ours]{\includegraphics[width=0.23\textwidth]{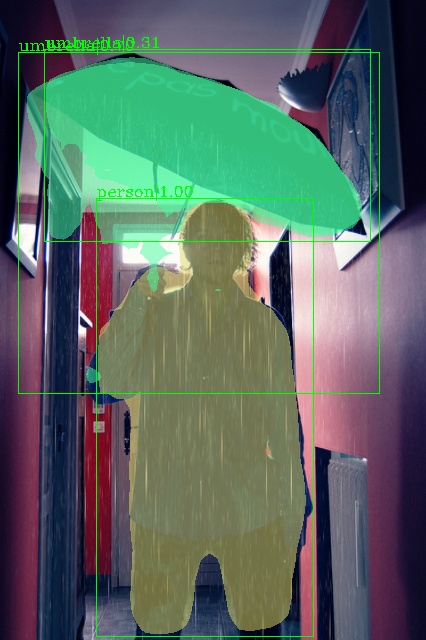}}\\
  \caption{Visualization of COCO dataset predicted results}
  \label{fig:fig6}
\end{figure*}

\subsection{Experimental results on COCO dataset}
 Compared with the Cityscapes dataset, the COCO dataset contains objects of more classes and scales, so it is more challenging for our algorithm. Note that we set the loss balance weight coefficient $\alpha$ to 0.1 experimentally on COCO dataset. The experimental results are shown in Table~\ref{tab:my-table9}. Notably, compared with baseline, our method has effective gains in both instance segmentation and target detection tasks. Moreover, in Table~\ref{tab:my-table10} we compare our results with state-of-the-art counterparts on COCO test-dev dataset, and it can be seen from the experimental results that we have obtained comparable performance. 

\subsection{Qualitative Evaluation on COCO dataset}
In order to analyze the performance of our algorithm more intuitively, we show some concrete examples of our method and Mask R-CNN baseline in Fig.~\ref{fig:fig6}. It can be seen from the figure that, compared with the baseline, our method can obtain a segmentation effect that fits the contour of the object better, which is thanks to our newly added keypoint detection and auxiliary task and effective fusion strategy.

\section{Discussion and Future work}
\label{futurework}
In this section, we will analyze the effectiveness of our method from the view of multi-task joint training and introduce our future work.

In fact, in Mask R-CNN, it has also carried out multi-task joint training of object detection, keypoint estimation, and instance segmentation on the person category in the COCO dataset. However, the experimental results show that only the performance of the keypoint estimation task is slightly improved, whereas both the performance of the detection and segmentation tasks is degraded. We deduce that this is because in the task of human pose estimation, the labels of keypoints are located within the human body, and the overall information of the instances captured by the segmentation task can help locate these keypoints. But for the object detection task including positioning and classification, and the segmentation task composed of pixel-wise classification, the keypoint information does not help, but after we set the weight balance parameters of the keypoint detection task ${\alpha = 0.5}$, we can see that the detection performance of these three tasks has been greatly improved. From the above experimental results, we can infer that the performance of the model can be promoted by exploring better balance coefficients in the process of multi-task joint training. 

Moreover, in order to verify the effectiveness of our method more intuitively, we visualize the final feature map obtained by the keypoint head and the corresponding feature map obtained by the mask head. Specifically, we normalize the two feature maps respectively and generate corresponding heat maps, some examples are demonstrated in Fig.~\ref{fig:fig5} (the darker pixels in the heat map represent higher scores). From these examples we can infer that the keypoint output feature map obtained by channel-wise addition can not only learn the boundary information of the object, but also get the overall background information. Therefore, the final output feature of the mask branch obtained by fusing the two tasks is equivalent to segment the object by combining the background and foreground features. Besides, in the joint training, we can get the edge point information of the target through the keypoint estimation task. Actually, the target can be segmented by only using the contour information mapped by these edge points. When $k = 100$, we can get AP value of 12.5 on Cityscapes val subset.

In addition, our work still has some problems that need to be solved. First, compared to the original Mask R-CNN, the newly added keypoint branch has more parameters, which will also increase the computational burden, therefore, in the future we need to explore more lightweight designs to optimize our methods. And as mentioned above, our model has a strong sensitivity to the setting of the balance coefficient $\alpha$ in the multi-task joint training with different datasets, which is also a direction that needs further research.

\section{Conclusion}
\label{Conclusion}
In this paper, we propose Mask Point R-CNN for instance segmentation, which can enhance the consistency between segmentation prediction results and groundtruths masks by adding edge aggregation point auxiliary detection tasks. The experiments on the Cityscapes dataset and COCO dataset have proved the effectiveness of our method. We hope that our method can have some inspiration for other instance segmentation methods.



\bibliographystyle{elsarticle-num} 
\bibliography{maskpoint}
\biboptions{numbers,sort&compress}




\end{document}